\documentclass{article}

\usepackage[preprint]{neurips_2025}

\usepackage[utf8]{inputenc} % allow utf-8 input
\usepackage[T1]{fontenc}    % use 8-bit T1 fonts
\usepackage{hyperref}       % hyperlinks
\usepackage{url}            % simple URL typesetting
\usepackage{booktabs}       % professional-quality tables
\usepackage{amsfonts}       % blackboard math symbols
\usepackage{nicefrac}       % compact symbols for 1/2, etc.
\usepackage{microtype}      % microtypography
\usepackage{xcolor}         % colors
\usepackage{graphicx}
\usepackage{amsmath}
\usepackage{amssymb}
\usepackage{amsthm}
\usepackage{subcaption}
\usepackage{bm}
\usepackage{threeparttable}
\usepackage{diagbox}
\usepackage{float}
\usepackage{multirow}

\usepackage{cases}
\usepackage{wrapfig}

\usepackage[capitalize]{cleveref}
\crefname{section}{Section}{Sections}
\Crefname{section}{Section}{Sections}
\Crefname{table}{Table}{Tables}
\crefname{table}{Tab.}{Tabs.}
\usepackage{xcolor}
\definecolor{newgreen}{RGB}{0,176,80}
\definecolor{newblue}{RGB}{0,176,240}
\newcommand{\dprime}{{\prime\prime}}

\title{A Self-Ensemble Inspired Approach for Effective Training of Binary-Weight Spiking Neural Networks}

\author{\hspace{-1.5em}
  Qingyan Meng$^1$ 
  \
  Mingqing Xiao$^2$ 
  \
  Zhengyu Ma$^1$  
  \
  Huihui Zhou$^1$  
  \
  Yonghong Tian$^{1,3}$  
  \
  Zhouchen Lin$^3$ 
\\
\vspace{1pt}
$^1$Pengcheng Laboratory  \ \
$^2$Microsoft Research Asia – Shanghai \ \
$^3$Peking University\\
}

\begin{document}

\maketitle

\begin{abstract}
  Spiking Neural Networks (SNNs) are a promising approach to low-power applications on neuromorphic hardware due to their energy efficiency. However, training SNNs is challenging because of the non-differentiable spike generation function. To address this issue, the commonly used approach is to adopt the backpropagation through time framework, while assigning the gradient of the non-differentiable function with some surrogates. Similarly, Binary Neural Networks (BNNs) also face the non-differentiability problem and rely on approximating gradients. However, the deep relationship between these two fields and how their training techniques can benefit each other has not been systematically researched. Furthermore, training binary-weight SNNs is even more difficult. In this work, we present a novel perspective on the dynamics of SNNs and their close connection to BNNs through an analysis of the backpropagation process. We demonstrate that training a feedforward SNN can be viewed as training a self-ensemble of a binary-activation neural network with noise injection. Drawing from this new understanding of SNN dynamics, we introduce the \underline{S}elf-\underline{E}nsemble \underline{I}nspired training method for (\underline{B}inary-\underline{W}eight) \underline{SNN}s (SEI-BWSNN), which achieves high-performance results with low latency even for the case of the 1-bit weights. Specifically, we leverage a structure of multiple shortcuts and a knowledge distillation-based training technique to improve the training of (binary-weight) SNNs. Notably, by binarizing MLP layers in a Transformer architecture, our approach achieves 82.52\% accuracy on ImageNet with only 2 time steps, indicating the effectiveness of our methodology and the potential of binary-weight SNNs.

  % leads to \textbf{binary-weight} SNNs that achieve an accuracy of 82.52\% on ImageNet with only 2 time steps, indicating the effectiveness of our methodology and the potential of binary-weight SNNs.
\end{abstract}

\section{Introduction}

Spiking Neural Networks (SNNs) have emerged as the third generation of neural networks \citep{maass1997networks}, drawing inspiration from the communication mechanisms of biological neurons that rely on spikes to transmit information. The promise of SNNs lies in their energy efficiency when implemented on neuromorphic hardware \citep{schuman2017survey}, which makes them particularly well-suited for low-power applications. In contrast, the power consumption of deep Artificial Neural Networks (ANNs) is substantial.

Despite the potential of SNNs, their supervised training presents a challenge due to the non-differentiability of the spike generation process. To address this issue, one of the most effective training methods currently available is the Surrogate Gradient (SG) \citep{neftci2019surrogate,cramer2022surrogate} method. This method employs the framework of backpropagation through time (BPTT) \citep{werbos1990backpropagation} and uses surrogate function to approximate the gradient of the non-differentiable spike generation function, yielding relatively satisfactory accuracy results with low latency (\emph{i.e.}, short spike train lengths). In parallel, Binary Neural Networks (BNNs) \citep{courbariaux2016binarized,hubara2016binarized} have emerged as a research topic where weights and activations are binarized to 1 bit. Despite being an orthogonal research area, BNN training also faces the challenge of non-differentiable step functions \citep{rastegari2016xnor}, much like SNNs, and also relies on approximating gradients of the step functions with various surrogates. While these two fields share similar training problems, %how their training techniques can mutually benefit each other's development remains unclear.
how their training techniques can benefit each other's development has not been systematically researched.

Furthermore, there has been limited works on binary-weight spiking neural networks (BWSNNs), without yielding satisfactory results. Although not being fully researched, the topic of BWSNN holds great potential. Implementing binary weights in neuromorphic chips can significantly reduce memory usage, which is particularly important as these chips often have limited storage resources. Additionally, the binary operations executed by BWSNNs can be more efficiently processed on these chips. However, training a BWSNN poses a greater challenge than a full-precision SNN due to the much more chaotic loss landscape.

In this study, we present a novel approach to SNN training by providing a new perspective on their dynamics and their connection to BNNs. Our proposed method is termed Self-Ensemble Inspired training method for BWSNNs (SEI-BWSNN). Through analyzing the backpropagation process, we discover that training a feedforward BWSNN can be viewed as training a self-ensemble of a BNN with noise injection, as both follow similar weight update paths. Building on this new perspective, we improve BWSNN performance by enhancing the BNN sub-network. The improvement is made by employing a structure of multiple shortcuts and an optimization process based on knowledge distillation \citep{hinton2015distilling}. These simple yet effective methods achieve good performance, bolstering the validity of our new perspective. Additionally, we develop a training pipeline that results in high-performance networks with low latency.
Formally, our key contributions include:
\begin{itemize}
	\item[1.]	We present a novel perspective on SNN dynamics. Our analysis of backpropagation reveals that a feedforward SNN can be regarded as a self-ensemble of a binary network with noise injection. 
    %This conceptual framework provides novel insights into BPTT-based training processes, offering a foundation for developing more effective SNN training approaches.
    We posit that this perspective advances the community's understanding of BPTT-based training and consequently improves SNN training methodologies.
    %We posit that this conceptual perspective helps the community build a new understanding on the BPTT-based training process and consequently improve SNN training.	
	\item[2.] Based on the conceptual perspective on SNN dynamics, we pave a new path for training SNNs by leveraging BNN training techniques. The proposed methods, including a multiple-shortcut structure and a knowledge distillation process, are simple yet effective in enhancing SNN performance.	

	\item[3.] We provide a strong baseline for training binary-weight SNNs (BWSNNs) with low latency. Notably, the obtained model achieves an accuracy of 82.52\% on ImageNet with only 2 time steps.
    % The obtained BWSNN models achieve accuracies of 95.04\%, 74.45\%, and 62.08\% on CIFAR-10, CIFAR-100, and ImageNet, respectively, with only 6 time steps, and achieve an accuracy of 82.40\% on DVS-CIFAR10 with only 10 time steps. 
    Our BWSNN models outperform many state-of-the-art models that rely on full-precision weights. The results highlight the potential of BWSNN as a promising approach for high-performance artificial intelligence with ultra-low power consumption.
\end{itemize}

\vspace{-5pt}
\section{Related Work}
\vspace{-5pt}
\paragraph{SNN Training Methods}
Supervised learning for SNNs is typically conducted through one of three main directions. The first involves the ANN-to-SNN conversion method \citep{yan2021near, deng2021optimal, sengupta2019going, rueckauer2017conversion, han2020rmp, han2020deep, ding2021optimal, li2021free, meng2022ann, bu2022optimal,wang2023toward}, which establishes a connection between the functionalities of an SNN and some corresponding ANN. Based on this connection, SNN parameters are directly copied from the trained associated ANN. The second direction involves finding an equivalence between the firing rates or first spike times of SNNs and some ANN-like differentiable mappings \citep{zhou2019temporal, xiao2021ide, meng2022training, thiele2019spikegrad, wu2021training, wu2021tandem, xiao2022spide, yang2022training, yang2023lc}. Gradients calculated from these mappings are then used to optimize the SNNs. Despite good performance from recent works \citep{bu2022optimal,meng2022training,huang2024towards}, methods from the above two directions typically require high latency (\emph{i.e.}, a large number of time steps), leading to high energy consumption. To achieve high performance with low latency, the third direction borrows the BPTT framework and assigns surrogate gradients (SG) to the non-differentiable spike generation to enable valid backpropagation \citep{zenke2021remarkable, neftci2019surrogate, wu2018spatio, wu2019direct, shrestha2018slayer, ma2023exploiting}. Under the BPTT framework, many effective techniques have been proposed to improve performance \citep{zheng2020going, fang2021sew, li2021differentiable, guo2022recdis, deng2022temporal, fang2021incorporating, li2022neuromorphic, zhang2021rectified, chowdhurytowards, guo2022reducing, deng2022temporal, yan2024sampling}, or reduce computational costs \citep{meng2022towards, xiao2022online, bellec2020solution, bohnstingl2022online, yin2021accurate, zhang2023self}. Among the BPTT-based methods, SLTT \citep{meng2022towards} reduces the computational complexity of the BPTT framework significantly by ignoring some components in gradient calculation while maintaining the same level of performance. This method inspired our rethinking of SNN dynamics. 
In our work, we focus on the BPTT framework that has achieved high performance with low latency on both static and neuromorphic datasets.
We design learning techniques based on our new perspective on the SNN dynamics.

\paragraph{Binarization and Ultra-Low-Bit Quantization for Neural Networks} Some pioneering works \citep{courbariaux2016binarized,hubara2016binarized} have demonstrated the feasibility of training Binary Neural Networks (BNNs), which are ANNs with binarized weights and activations. Many attempts to improve performance have focused on network structure design, such as introducing real-valued skip connections \citep{liu2018bi,liu2020reactnet}, channel re-scaling \citep{martinez2020training,zhang2022pokebnn}, non-linearities \citep{bulat2019improved,liu2020reactnet}, and network ensembles \citep{zhu2019binary}. Another comprehensive line of works is the multi-stage training strategy \citep{bulat2019improved,liu2021adam,zhuang2018towards,yang2019synetgy}, where training is conducted step-by-step for (unquantized models,) binary-activation-only models, and BNN models. Meanwhile, many BNN training methods \citep{liu2018bi,liu2020reactnet,martinez2020training,zhang2022pokebnn} heavily depends on knowledge distillation \citep{hinton2015distilling} to provide a more fine-grained supervision signal from teacher models. Recently, BitNet \cite{wang2023bitnet,ma2024era} employs binary (or ternary) weight quantization for large language models (LLMs) and exhibits a scaling law, showing the potential of ultra-low-bit quantization in the LLM era.
Regarding SNNs, weight binarization and ultra-low-bit quantization have been explored to reduce energy consumption further \cite{quantizedspikedriventransformer,binaryeventdrivenspikingtransformer,kheradpisheh2022bs4nn,q-snns}.
Many works focus on the hardware implementation aspect and show experiments on simple datasets like MNIST
\citep{koo2020sbsnn,srinivasan2019restocnet,hu2021quantized,qiao2021direct,wang2020deep,EsserMACAABMMBN16,wei2021binarized}. On the algorithm aspect, 
QP-SNNs \cite{qp-snns} employs a weight rescaling strategy to maximize bit-width efficiency, demonstrating successful training with 2-bit weights on various small-scale benchmarks. Lu et al. \cite{lu2020exploring} use the ANN-to-SNN conversion method to convert a BNN to a BWSNN and achieve satisfactory performance on ImageNet, but at the cost of high latency. AGMM \cite{AGMM} proposes an adaptive gradient modulation mechanism binary-weight SNNs and achieves an accuracy of over 60\% on ImageNet with low latency.
In this work, we propose a training method for BWSNNs inspired by our new perspective that the BWSNN dynamics can be treated as a BNN self-ensemble, enabling us to achieve SOTA performance with low latency.

%\textbf{Ultra-Low-Bit Quantization in Spiking Neural Networks.} 
%SNNs are event-driven, utilizing additive operations between transmitted spikes and synaptic weights, which endows them with low power consumption and high computational efficiency. Combining weight quantization further amplifies these intrinsic advantages.  The Q-SNNs \cite{q-snns} binarize synaptic weights into $\{+1, -1\}$, and introduce a loss function to constrain neuronal firing rates approaching 0.5, and further reduce memory footprint by implementing 8-bit quantization on membrane potentials. AGMM \cite{AGMM} proposes an adaptive gradient modulation to ensure the performance after quantization. In addition to the network weights, BESTformer \cite{binaryeventdrivenspikingtransformer} further quantizes the self-attention map to 1-bit, with the resultant performance degradation being compensated by the proposed Coupled Information Enhancement strategy. While existing methods attain efficient computation and model compression via integrating SNNs with low-bit quantization, they persistently suffer from performance degradation.

\section{Preliminaries}
\label{sec:preliminary}

\begin{figure*}[t]
	%\centering
	\begin{subfigure}{0.33\linewidth}
		\includegraphics[height=0.74\columnwidth]{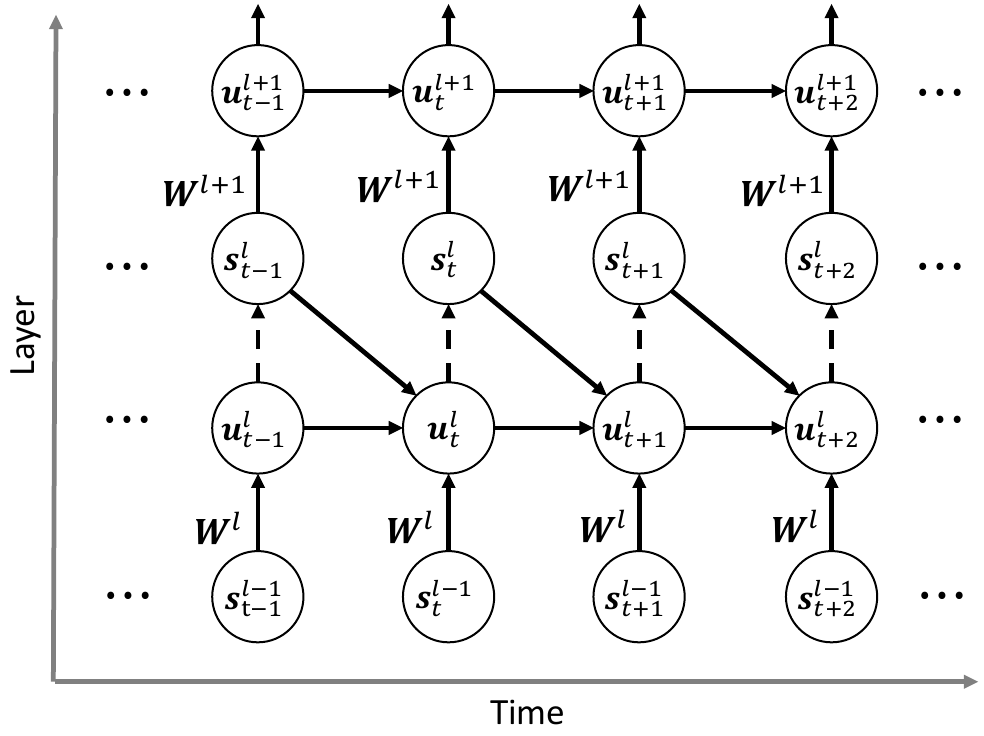}
		\caption{ SNN forward} 
		\label{fig:BP-forward}
	\end{subfigure}
	\begin{subfigure}{0.33\linewidth}
		\includegraphics[height=0.74\columnwidth]{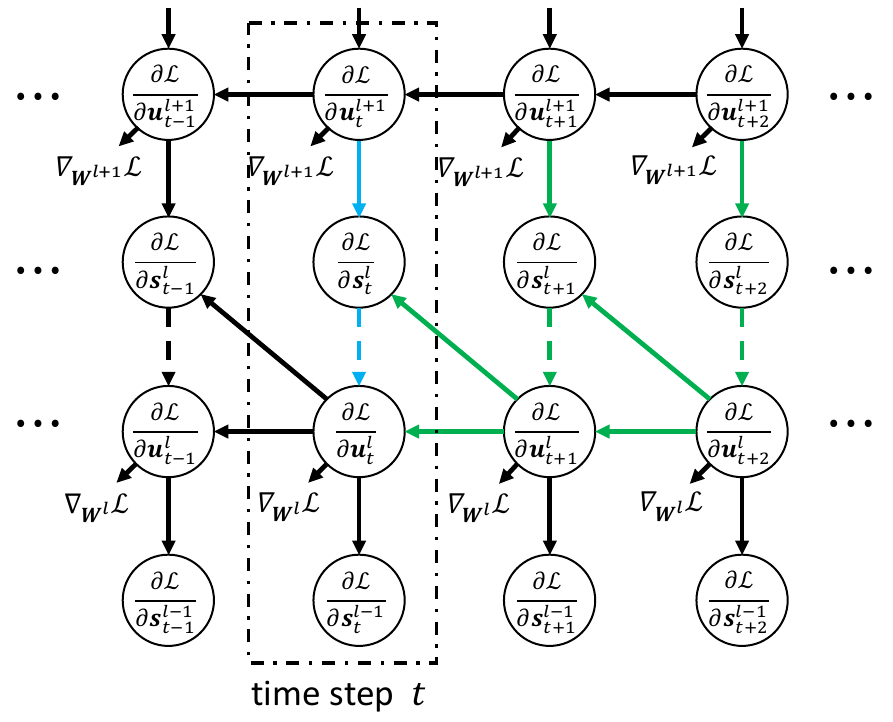}
		\caption{ SNN backward: SG}  
		\label{fig:BP-SG}
	\end{subfigure}
	\begin{subfigure}{0.33\linewidth}
		\includegraphics[height=0.74\columnwidth]{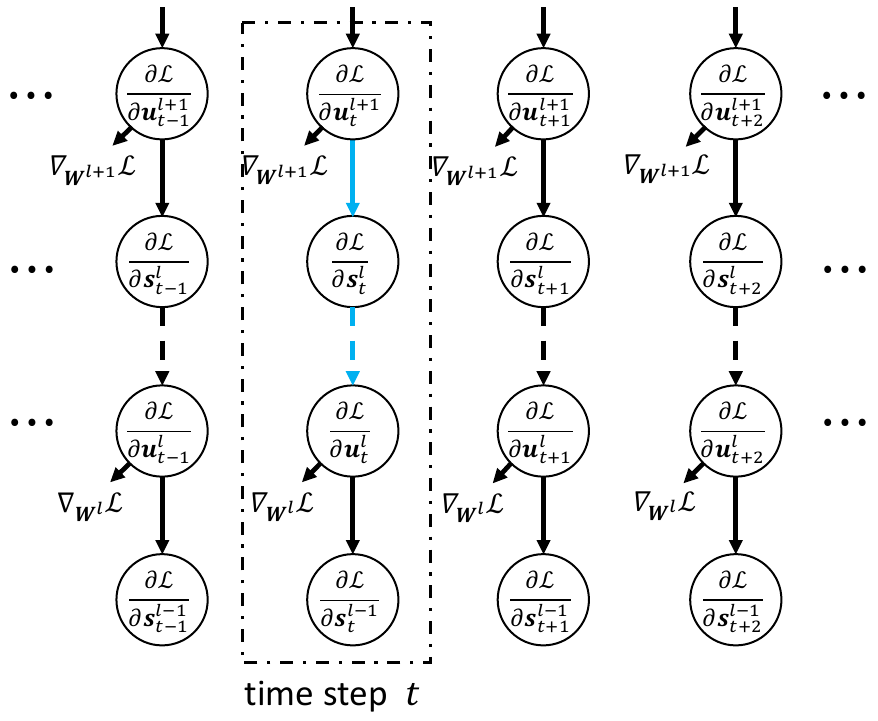}
		\caption{ SNN backward: SLTT} 
		\label{fig:BP-SLTT}
	\end{subfigure}
	\caption{\normalsize The forward and backward computation of a feedforward SNN. Dashed arrows represent the non-differentiable spike generation functions. $x^{l}_t$ in the figures means $x^l[t]$ in \cref{eqn:feedforward,eqn:bptt,eqn:bptt0,eqn:sltt} where `$x$' can be $\mathbf{u,s},$ or $\mathbf{W}$.}
\end{figure*}

\paragraph{Spiking Neural Networks} SNNs mimic the behavior of biological neural networks, which transmit information through binary spike trains. In an SNN, each neuron-$i$ integrates input spikes into its membrane potential $u_i$ and generates a spike once $u_i$ reaches a threshold. The commonly used leaky integrate and fire (LIF) model is adopted in this work to describe the dynamics of $u_i$:
\begin{subequations}  \label{eqn:lif}
	\begin{align}
	&u_i[t]=\lambda \cdot v_i[t-1] + \sum_{j}w_{ij} s_j[t] + b, \label{eqn:lif1}\\
	&s_{i}[t]=H\left(u_i[t]-V_{th}\right), \label{eqn:lif2}\\[2pt]
	&v_i[t]=u_i[t]-V_{th} \cdot s_{i}[t], \label{eqn:lif3}
	\end{align}
\end{subequations}
where $t\in\{1,2,\cdots,T\}$ is the time step index, $\lambda \in (0,1)$ is a leaky term which is set as $0.1$ throughout the paper, $v_i[t]$ is the intermediate state of the membrane potential with $v_i[0]=0$, $s_{j}[t] \in \{0,1\}$ is the input spike from neuron-$j$,
$w_{ij}$ is the weight from neuron-$j$ to the target neuron-$i$, $b$ is a bias term, $s_{i}[t] \in \{0,1\}$ is the output spike of the target neuron-$i$, $H(\cdot)$ is the Heaviside step function, and $V_{th}$ is the spike threshold. \cref{eqn:lif1,eqn:lif2,eqn:lif3}
represent the potential integration, spike generation, and potential resetting processes, respectively. For simplicity, we ignore the bias term $b$ in the following. 
%From \cref{eqn:lif3}, we can get 
%\begin{equation} \label{eqn:lif3_}
%    v_i[t-1]=u_i[t-1]-V_{th}\cdot s_i[t-1].
%\end{equation}
%By substituting \cref{eqn:lif3_} into \cref{eqn:lif1}, 
A feedforward SNN, which consists of layers of LIF neurons, can be characterized as:  
\begin{equation}  \label{eqn:feedforward}
\mathbf{u}^{l}[t]=\mathbf{n}^{l}[t-1]+ \mathbf{W}^{l} \mathbf{s}^{l-1}[t],
\end{equation}
where
\begin{equation} 
\mathbf{n}^{l}[t-1]=\lambda(\mathbf{u}^{l}[t-1] -V_{t h} \mathbf{s}^{l}[t-1]),
\end{equation}
$l=1,2,\cdots,L$ is the layer index, and $\mathbf{W}^{l}$ is the weight from layer $l-1$ to $l$. The computational graph is shown in \cref{fig:BP-forward}. % The term $\mathbf{n}^{l}[t]$ can be treated as ``noise'' in our following analysis.
While the term $\mathbf{n}^{l}[t-1]$ is obtained deterministically by
the LIF neuronal model, it will be treated as a systematic “noise” in our following analysis, as shown in \cref{sec:method}.
SNNs leverage additive operations between transmitted spikes and synaptic weights. Using binary weights in SNN can further reduces accumulation operations to XNOR with popcount operations, which are more energy efficient and can be supported in neuormorphic chips, showing great potential.

\paragraph{The Surrogate Gradient Method For Training SNNs.}

To learn the weights in an SNN, the BPTT framework directly calculates the gradient %with respect to weights $\mathbf{W}^{l}$ 
based on the computational graph \citep{meng2022towards}:
\begin{equation} \label{eqn:bptt0}
\begin{aligned}
\frac{\partial \mathcal{L}}{\partial \mathbf{W}^{l}} 
=&
\sum_{t=1}^{T} 
\mathbf{s}^{l-1}[t]
\frac{\partial \mathcal{L}}{\partial \mathbf{u}^{l}[t]},
\end{aligned}
\end{equation}
where $\mathcal{L}$ is the loss function, and 
\begin{equation}  \label{eqn:bptt}
\begin{aligned}
\frac{\partial \mathcal{L}}{\partial \mathbf{u}^{l}[t]}
%*
=   \textcolor{newblue}{
	\frac{\partial \mathcal{L}}{\partial \mathbf{u}^{l+1}[t]}
	%*
	\frac{\partial \mathbf{u}^{l+1}[t]}{\partial \mathbf{s}^{l}[t]}
	%*
	\frac{\partial \mathbf{s}^{l}[t]}{\partial \mathbf{u}^{l}[t]}} 
+  \textcolor{newgreen}{
	\sum_{t^\prime=t+1}^{T}
	%*
	\frac{\partial \mathcal{L}}{\partial \mathbf{u}^{l+1}[t^\prime]}
	%*
	\frac{\partial \mathbf{u}^{l+1}[t^\prime]}{\partial \mathbf{s}^{l}[t^\prime]}
	%*
	\frac{\partial \mathbf{s}^{l}[t^\prime]}{\partial \mathbf{u}^{l}[t^\prime]}
	%*
	\prod_{t^{\dprime}=1}^{t^\prime - t}
	%*
	\mathbf{\epsilon}^{l}[t^\prime-t^\dprime]
}
%*
,
\end{aligned}
\end{equation}
with the term $\mathbf{\epsilon}^{l}[t]$ defined as
\begin{equation}
\textcolor{black}{\mathbf{\epsilon}^{l}[t]} \triangleq
\frac{\partial \mathbf{u}^{l}[t+1]}{\partial \mathbf{u}^{l}[t]}
%*
+\frac{\partial \mathbf{u}^{l}[t+1]}{\partial \mathbf{s}^{l}[t]}
%*
\frac{\partial \mathbf{s}^{l}[t]}{\partial \mathbf{u}^{l}[t]}.
\end{equation}
However, the non-differentiability of \cref{eqn:lif2} results in a zero value (almost surely) for $\frac{\partial \mathbf{s}^{l}[t] }{\partial \mathbf{u}^{l}[t]}$ in \cref{eqn:bptt}, which prevents the update of weight $\mathbf{W}^{l}$. To overcome this issue, the surrogate gradient (SG) method \citep{neftci2019surrogate} approximates $\frac{\partial \mathbf{s}_i^{l}[t] }{\partial \mathbf{u}_i^{l}[t]}$ with some well-behaved surrogate, such as the triangle function \citep{deng2022temporal}
\begin{equation} \label{eqn:triangle} 
\frac{\partial s}{\partial u}=\frac{1}{\gamma^2} \max \left(0, \gamma-\left|u-V_{t h}\right|\right),
\end{equation}
where $\gamma$ is a hyperparameter often set as $V_{th}$.

\paragraph{Binary Neural Netwotorks.}
Unlike SNNs, BNNs are a kind of artificial neural networks that do not have time dimension.
In a BNN, both activations and weights of the convolutional layers are binarized using the sign function such that
\begin{equation} \label{eqn:bnn}
x_b=\left\{\begin{array}{l}
+1, \text { if } x_r>0 \\
-1, \text { if } x_r \leq 0
\end{array}\right.
, \
w_b=\left\{\begin{array}{l}
+a, \text { if } w_r>0 \\[3pt]
-a, \text { if } w_r \leq 0
\end{array}\right.
,
\end{equation}
where $x_r$ and $w_r$ indicate the real activation and weight before binarization, respectively, maintained in the training process, and $a$ is a real-valued scaling factor. To minimize the difference between the binary and full-precision weights, $a$ is chosen as  $\frac{\left\|\mathbf{W}_r\right\|_{1}}{n}$ \citep{rastegari2016xnor}, where $n$ is the total number of parameters in the convolution kernels $\mathbf{W}_r$. 
BNNs provide a remarkable reduction in memory usage, up to $32\times$, and significantly faster computation speeds compared to their real-valued counterparts.
Similar to SNN training, BNN training is also challenging due to the non-differentiable sign function used in binarization. As a result, surrogate functions such as the straight-through estimator \citep{bengio2013estimating} are commonly used to approximate the derivatives of the sign functions. If only the activations are binarized, the resulting network is a binary-activation neural network (BANN). %, as discussed in \cref{sec:method}.

\vspace{-5pt}
\section{Methodology}
\label{sec:method}
\vspace{-5pt}
\subsection{SNN as Self-Ensemble of A Binary-Activation Neural Network}
\label{sec:rethink}

In this subsection, we provide a novel interpretation of the forward computation of an SNN by treating it as a self-ensemble of a BANN, viewed through the lens of BPTT-based training. Building upon this observation, we introduce a set of techniques to improve the performance of (binary-weight) SNNs, which are discussed in detail in \cref{sec:improve-snn,sec:pipline}.

The SG method is a BPTT-based method for training SNNs.
It involves backpropagating error signals through both the layer-wise spatial domain and the time-wise temporal domain, as depicted in \cref{fig:BP-SG}. For each time step $t$, the calculation for $\frac{\partial \mathcal{L}}{\partial \mathbf{u}^{l}[t]}$ in \cref{eqn:bptt} requires information from both the current time step $t$ and future time steps, which are shown as blue and green paths in \cref{fig:BP-SG}, respectively.
Interestingly, Meng et al. \cite{meng2022towards} observe that ignoring the future time steps during the backward process has little effect on the calculated value of $\frac{\partial \mathcal{L}}{\partial \mathbf{u}^{l}[t]}$ or the gradient of weights $\frac{\partial \mathcal{L}}{\partial \mathbf{W}^{l}}$. With the observation, their proposed spatial learning through time (SLTT) method updates the weights by
\begin{equation} \label{eqn:sltt}
\begin{aligned}
\frac{\partial \mathcal{L}}{\partial \mathbf{W}^{l}}
%*
&=\sum_{t=1}^{T} 
\mathbf{s}^{l-1}[t]
\frac{\partial \mathcal{L}}{\partial \mathbf{u}^{l}[t]}
, \\[1pt]
\frac{\partial \mathcal{L}}{\partial \mathbf{u}^{l}[t]} &=
\textcolor{newblue}{
	\frac{\partial \mathcal{L}}{\partial \mathbf{u}^{l+1}[t]}
	%*
	\frac{\partial \mathbf{u}^{l+1}[t]}{\partial \mathbf{s}^{l}[t]}
	%*
	\frac{\partial \mathbf{s}^{l}[t]}{\partial \mathbf{u}^{l}[t]}} 
%*
,
\end{aligned}
\end{equation}
in which the temporal information (green parts in \cref{eqn:bptt} and \cref{fig:BP-SG}) is ignored. The backpropagation process of SLTT is illustrated in \cref{fig:BP-SLTT}. According to theoretical analysis and experimental proof in Meng et al. \cite{meng2022towards}, SLTT %(\cref{eqn:sltt}) 
and SG %(\cref{eqn:bptt0,eqn:bptt}) 
have similar gradient descent directions and performances under different surrogate functions and hyperparameters.

The SLTT learning rule offers a fresh perspective on the dynamics of SNNs. As the backward process in SLTT is decoupled across time steps, the corresponding forward process can be viewed as a self-ensemble of a BANN with data-dependent ``noise'' injected into the pre-activations, as shown in \cref{fig:ensemble}. In this self-ensemble, all sub-networks have the same weights but varied injected ``noise'' generated by the LIF model.
In each weak sub-network, $\mathbf{u}^l[t]$ represents the pre-activation, $\mathbf{s}^l[t]$ represents the post-activation, and binarization is defined in \cref{eqn:lif2}. This binarization method is also present in some BNN models \citep{wang2020sparsity,tu2022adabin}, in contrast to the standard binarization method described in the left-hand side of \cref{eqn:bnn}.
As SG and SLTT exhibit similar behaviors, SNN dynamics and the BANN self-ensemble enjoy similar performances following gradient-based training.

The self-ensemble with injected ``noise'' is expected to improve robustness and accuracy compared to the single weak BANN sub-network \citep{noh2017regularizing,liu2018towards,wang2019resnets,zhu2019binary}, especially when the SNN is with high latency. According to the proposed perspective, higher latency means that a larger number of weak BANN learners are involved in the ensemble, resulting in better and more robust performance. The effect of the self-ensemble can be one reason why high-latency (large $T$) SNNs perform better than their low-latency counterparts \citep{chowdhurytowards}, even for static datasets. On the contrary, based on our analysis, the use of temporal information may be not the main reason why high-latency SNNs perform better.

% \begin{figure}[t]
% 	\centering
% 	\includegraphics[width=0.5\columnwidth]{figures/ensemble.pdf} 

% 	\caption{\normalsize The illustration of our perspective on the SNN dynamics. The communication through time can be decoupled from the computational graph and treated as data-dependent ``noise''. $\mathbf{n}_t^l$ represents the ``noise'' injected, and numerically equals to $\lambda(\mathbf{u}^{l}[t-1] -V_{t h} \mathbf{s}^{l}[t-1])$.}

% \end{figure}

\vspace{-5pt}
\subsection{Improving SNN Performance with BNN Training Techniques}
\label{sec:improve-snn}
\vspace{-5pt}
Regarding a feedforward SNN as a BANN self-ensemble, it is intuitive to consider enhancing the SNN performance by improving the constituent BANN sub-network (weak learner). Fortunately, there exists a substantial body of literature on BNNs, including BANNs, that can furnish useful guidance on optimizing SNN performance and identifying effective methodologies for training binary-weight SNNs which can be viewed as BNN self-ensembles.

\begin{figure}[t] 
	%\centering
    \begin{subfigure}{0.33\linewidth}
		\centering
		\includegraphics[height=0.8\columnwidth]{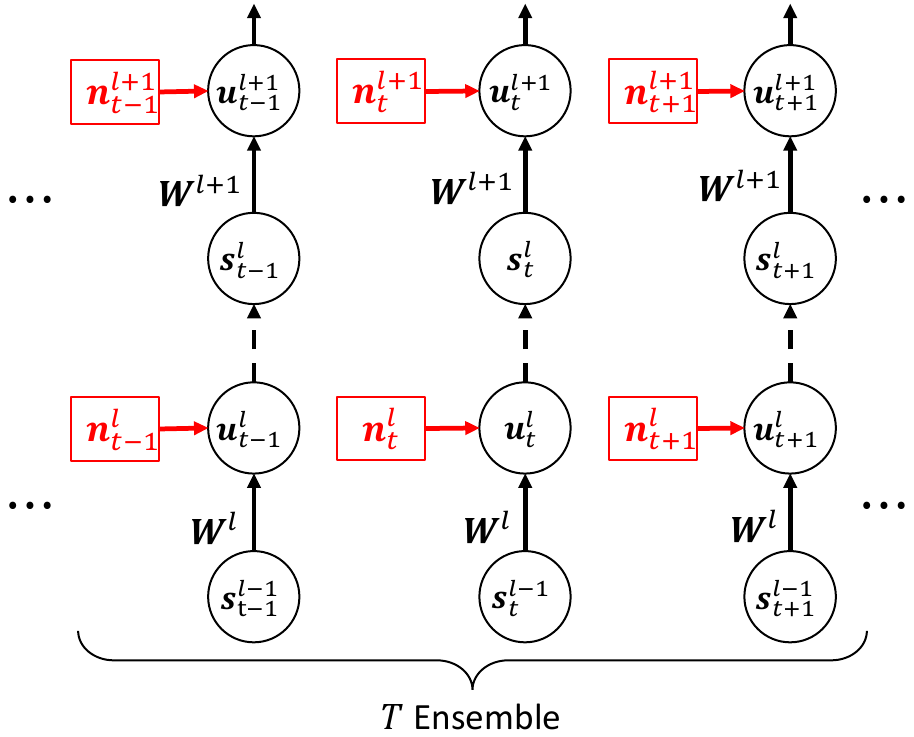}
		\caption{} 
		\label{fig:ensemble}
	\end{subfigure}
	\begin{subfigure}{0.33\linewidth}
		\centering
		\includegraphics[height=0.8\columnwidth]{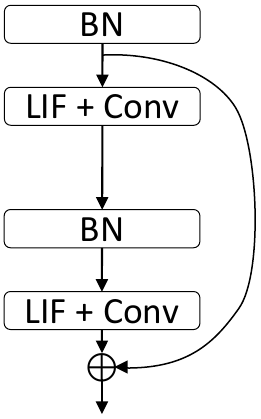}
		\caption{} 
		\label{fig:block}
	\end{subfigure}
	\begin{subfigure}{0.33\linewidth}
		\centering
		\includegraphics[height=0.8\columnwidth]{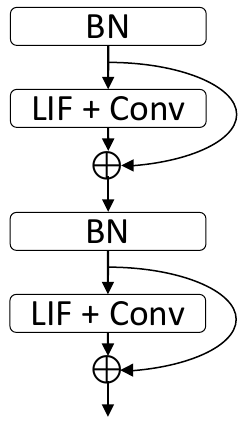}
		\caption{}  
		\label{fig:double-block}
	\end{subfigure}

	\caption{(a) The illustration of our perspective on the SNN dynamics. The communication through time can be decoupled from the computational graph and treated as data-dependent ``noise''. $\mathbf{n}_t^l$ represents the ``noise'' injected, and numerically equals to $\lambda(\mathbf{u}^{l}[t-1] -V_{t h} \mathbf{s}^{l}[t-1])$. (b) The original basic block structure in a spiking ResNet of the pre-activation version \citep{he2016identity}. (c) The proposed basic block structure with double skip connections bypassing the LIF neurons and the convolutions.}
	\label{fig:structure}

\end{figure}

First, a key technique used in many BNN architectures is the incorporation of multiple shortcuts bypassing the intractable binary activations and (1-bit) convolutions \citep{liu2018bi,liu2020reactnet,bethge2021meliusnet}. This technique has been shown to bring higher representational capability by keeping both real and binary activations. However, the SNN community has yet to discover this idea in the absence of our perspective on the SNN dynamics. We propose to use this idea to reconstruct the commonly used spiking ResNet architecture for better performance.
Unlike standard ResNet models \citep{he2016deep}, spiking ResNets have the same structures but with all ReLU activations replaced by LIF neurons. 
Regarding the network reconstruction, in the original pre-activation residual block \citep{he2016identity} in an SNN, multiple ``BN-LIF-Conv'' modes are used in the basic block or the bottleneck structures, respectively, with just one shortcut, as shown in \cref{fig:block}. We propose to introduce multiple shortcuts, one per mode, as shown in \cref{fig:double-block}. In such structures, all the ``LIF-Conv'' sub-modes are bypassed by the skip connections, mitigating information loss due to binary-output LIF neurons (and binarized convolutions).

Next, knowledge distillation (KD) techniques \citep{hinton2015distilling} are extensively used in recent BNN studies. 
Being guided by full-precision teacher networks, binary student networks can obtain more fine-grained supervision signals, leading to significantly improved performances. 
In this paper, with our new perspective on SNN dynamics, we introduce a simple yet effective KD technique, the KL-divergence loss \citep{liu2020reactnet,zhang2022pokebnn,tu2022adabin}, to improve SNN training. Specifically, the KL-divergence loss for BNNs is defined as:
\begin{equation} \label{eqn:loss}
\mathcal{L}_{\text {BNN }} =-\frac{1}{n} \sum_c \sum_{i=1}^n \rho_c^{\mathcal{R}}\left(x_i\right)  \log \left(\frac{\rho_c^{\mathcal{B}}\left(x_i\right)}{\rho_c^{\mathcal{R}}\left(x_i\right)}\right),
\end{equation}
where $x_i$ is the input data, $c$ represents classes, $n$ is the batch size, and $\rho_c^{\mathcal{R}}$ and $\rho_c^{\mathcal{B}}$ are the softmax outputs of the full-precision teacher net and binary student net, respectively. For SNNs with $T$ time steps, the loss is defined as
\begin{equation}  \label{eqn:snn-loss}
\mathcal{L}_{\text {SNN}}=-\frac{1}{nT} \sum_{t=1}^T  \sum_{i=1}^n \sum_c \rho_c^{\mathcal{A}}\left(X_i[t]\right)  \log \left(\frac{\rho_c^{\mathcal{S}}\left(X_i\right)[t]}{\rho_c^{\mathcal{A}}\left(X_i[t]\right)}\right),
\end{equation}
where $X_i$ is the time sequence input data with $X_i[t]$ being the data at the $t$-th time step, and $\rho_c^{\mathcal{A}}$ and $\rho_c^{\mathcal{S}}$ are the softmax outputs of the full-precision ANN teacher net and (binary) SNN student net, respectively. Note that the teacher network may have a different (typically more advanced) structure compared with the student network. Although some works \citep{yang2022training,kundu2021spike} on SNN training also employ ANN ``teachers'', they essentially rely on ANN-to-SNN conversion, where the ANN teachers have to share the same network structures as the SNN counterparts.

In the preceding discussion, we have only used two widely accepted concepts in the BNN community. Nevertheless, there are numerous other techniques that could be applied to SNN training without significant additional costs. With our novel perspective on SNN dynamics, we anticipate exploring additional opportunities to enhance SNN performance using various BNN techniques.

\subsection{Training Binary-Weight Spiking Neural Networks}
\label{sec:pipline}

Weight binarization is a critical factor for SNNs, as it enables substantial model compression, facilitating their deployment on neuromorphic chips with limited storage resources. For instance, Loihi \citep{davies2018loihi} offers approximately 16 MB of fixed memory to store weights, meanwhile accommodating any weight precision between one and nine bits. With 1-bit weight precision, larger-scale networks or lower memory consumption can be achieved. Moreover, using binary weights in SNN reduces accumulation operations to XNOR with popcount operations, which are more energy efficient and can be supported in neuormorphic chips.
However, existing approaches to BWSNN training have not yielded satisfactory results. By considering the BWSNN dynamics as a BNN self-ensemble, we propose the Self-Ensemble Inspired BWSNN training baseline (SEI-BWSNN) as follows:

\begin{itemize}

	\item \textbf{Network Structure}. Use the multiple-shortcut ResNet illustrated in \cref{fig:double-block}. The sequence of operations within a block should be arranged as ``BN-LIF-BinaryConv'' described in \cref{sec:improve-snn}. This ordering has been demonstrated to be optimal for BNNs \citep{martinez2020training}. The shortcut design is beneficial for training both binary weights and binary spike signals.

	\item \textbf{Loss Function}. Employ the KL-divergence loss for SNNs as defined in \cref{eqn:snn-loss}. For the ResNet-18 or ResNet-34 (with multiple shortcuts) BWSNN models, the teacher model can be chosen as ResNet-50 or Wide-ResNet \citep{zagoruyko2016wide} ANN models to provide better supervision.
	
	\item \textbf{Update of Binary Weights}. To obtain binary weights in the forward pass, use magnitude-aware binarization as presented in \cref{eqn:bnn}. In the backward pass, employ a simple surrogate \citep{liu2018bi,liu2020reactnet} such as
	\begin{equation} 
	\frac{\partial \mathcal{L}}{\partial w_r} = \frac{\partial \mathcal{L}}{\partial w_b} \cdot \frac{\partial w_b}{\partial w_r} \approx \frac{\partial \mathcal{L}}{\partial w_b} \cdot \mathbb{I}\left(\left|w_r\right|<1\right).
	\end{equation}
	to enable training. Here $w_r$ and $w_b$ are the weight values before and after binarization.

	\item \textbf{Two-Stage Training}. A two-stage optimization strategy for BNNs \citep{bulat2019improved} can be employed for SNNs. We can train an SNN with real-valued weights first, and then use the resulting model as the initialization to train the BWSNN model. Positive weight decay can be utilized in the first stage and set to 0 in the second stage. Such a weight decay scheme can balance the network stability and initialization dependency \citep{liu2021adam}.
\end{itemize}

By following the SEI-BWSNN training baseline described above, we can observe the superior performance of BWSNN models, which are very close to their non-binary SNN counterparts, as shown in \cref{sec:experiments}.

\vspace{-5pt}
\section{Experiments}
\label{sec:experiments}
\vspace{-5pt}
In this section, we evaluate the effectiveness of our proposed method on CIFAR-10/100 \citep{krizhevsky2009learning}, ImageNet \citep{deng2009imagenet}, DVS-Gesture \citep{amir2017low}, and DVS-CIFAR10 \citep{li2017cifar10}. We set the threshold $V_{th}$ to 1 and adopt the triangle surrogate (\cref{eqn:triangle}) for training. To binarize the weights, we follow the common practice \citep{liu2018bi,liu2020reactnet,tu2022adabin} of preserving full precision in the initial downsample layer and the last fully-connected layer. To conduct BP-based SNN training, we utilize the SLTT method, unless otherwise specified. Our code is available at  \url{https://github.com/qymeng94/SEI-BWSNN}.

\vspace{-5pt}
\subsection{Ablation Studies}

\vspace{-5pt}

We conduct ablation studies to evaluate the impact of the proposed multiple shortcuts structure and the SNN KL-divergence loss. For the baseline setting, we utilized standard spiking ResNet models (\cref{fig:block}) and employed the TET loss \citep{deng2022temporal} given by 
\begin{equation} \label{eqn:snn-ce}
\mathcal{L}=\frac{1}{nT} \sum_{t=1}^T \sum_{i=1}^n \ell_{CE}\left(o[t],y\right),
\end{equation}
where $\ell_{CE}$ is the cross-entropy function,  $y$ is the label, and $o[t]$ is the output of the SNN at the $t$-th time step. More specifically, $o[t]=\mathbf{W}^o\mathbf{s}^L[t]$, where $\mathbf{W}^o$ is the weight of the classification head, and the $c$-th element of $o[t]$ is identical to $\rho_c^{\mathcal{S}}\left(X_i\right)[t]$ in \cref{eqn:snn-loss}. In contrast to the SNN KL-divergence loss in \cref{eqn:snn-loss}, \cref{eqn:snn-ce} employs the widely used cross-entropy function in the summation. Thus, the TET loss is a suitable option for the baseline setting. As DVS-CIFAR10 is a neuromorphic dataset, we do not provide an ANN teacher model for it.

\begin{table}[t] \small
	\caption{ Accuracies (in percentage) of different combinations of the two proposed techniques. "BL" indicates the baseline setting, ``MS'' indicates the structure of multiple shortcuts (\cref{fig:double-block}), ``KL'' indicates the SNN KL-divergence loss (\cref{eqn:snn-loss}). The values of mean and standard deviation are calculated based on 3 runs of experiments. The values in brackets show the performance drop of 1-bit SNNs compared with their 32-bit SNN counterparts. }
	\hspace{-21pt}
	\label{table:ablation}
	\begin{threeparttable}
		\begin{tabular}{c|cccc}
            \toprule
            \multicolumn{5}{c}{\bf 32-bit weights} \\
			\toprule  
			Dataset  & BL & BL+MS & BL+KL & BL+MS+KL \\
			\midrule 
			{CIFAR-10}&  94.73 & 94.51 & 95.07 & \bf{95.22}\\
			{CIFAR-100}  &  76.46 & 76.94 & 76.66 & \bf{77.22} \\
			{ImageNet ($T=1$)}  &  {59.15} & {60.91} & {63.70} & {\bf{63.85}} \\
			{DVS-CIFAR10} &  81.6 & \bf{83.00} & / & / \\
			\bottomrule
            \multicolumn{5}{c}{\bf 1-bit weights} \\
            \toprule  
		  Dataset  & BL & BL+MS & BL+KL & BL+MS+KL \\
			\midrule 
			{CIFAR-10} & $94.26\pm0.04 (0.48)$ & $94.14\pm0.05  (0.38)$ & $94.72\pm0.15  (0.35)$ & $\bf{95.04}\pm0.07 (\bf{0.18})$\\
			{CIFAR-100} & $70.45\pm0.83  (6.02)$ & $74.08\pm0.08  (2.86)$ & $74.13\pm0.10  (\bf{2.53})$ & $\bf{74.45}\pm0.22 (2.78)$ \\
			{ImageNet ($T=1$)} & 53.20 & 54.87 & 57.31 & \bf{57.56} \\
			DVS-CIFAR10 & $80.20\pm0.26  (1.40)$ & $\bf{82.40\pm0.53  (0.60)}$ & / & / \\
			\bottomrule
		\end{tabular}
	\end{threeparttable}
    \vspace{-10pt}
\end{table}

 For CIFAR-10, CIFAR-100, and DVS-CIFAR10, we employ the ResNet-18 architecture and set the total number of time steps $T$ as 6, 6, 20, and 10, respectively. For ImageNet, we test the ResNet-34 architecture of the batchnorm-free version \citep{brock2021high,chen2021bnn} and set $T$ to be 1, considering limited resources. The two-stage training technique shown in \cref{sec:pipline} is applied to these three datasets, where we first obtain the trained SNN models with 32-bit weights, and then fine-tune them to obtain the BWSNN models with 1-bit weights. 
 
The results for both SNN and BWSNN are presented in \cref{table:ablation}, showing that the proposed techniques have positive effects, and that the combination of the techniques always yields the best performance. Although the shortcut structure does not outperform the baseline on CIFAR-10, the performance gap is small, and the shortcuts still provide benefits when combined with the SNN KL-divergence loss. The results further support the validity of the proposed new perspective on SNN dynamics and demonstrate the effectiveness of using this perspective to develop SNN training techniques.

\subsection{BWSNNs with One Time Step}
\vspace{-5pt}
In the case of 1 time step, a BWSNN is essentially a BNN itself, not a self-ensemble of the BNN. Since the self-ensemble mechanism is intended to improve generalization and performance, a multi-time-step BWSNN should outperform its 1-time-step counterpart. We have verified this on the static CIFAR-10, CIFAR-100, and ImageNet datasets with ResNet architectures, as shown in \cref{table:steps}. The results indicate that BWSNNs offer a performance benefit over BNNs. Moreover, when deployed on neuromorphic hardware, the energy consumption of BWSNNs is low due to the event-driven fashion and bit-wise operations in 1-bit convolutions. Overall, BWSNN training is a promising direction. We do not compare our 1-time-step BWSNN with other advanced BNN methods as they use operations that are not feasible in SNN implementation, such as complicated non-linearities (e.g., PReLU \citep{he2015delving} and its variants), real-valued multiplication, and Squeeze-and-Excitation block \citep{hu2018squeeze}.

\begin{minipage}{0.52\linewidth}
	%\centering
	\small
	\captionof{table}{Comparison between 6 and 1 time steps. The results for CIFAR-10/100 are based on 3 run of experiments. ``PG'' means performance gap between 6 and 1 time steps.}
	%\hspace{-3.5mm}
    \begin{threeparttable}
		\begin{tabular}{l|ccc}
			\toprule  
			Dataset  & $T=6$ & $T=1$ & PG \\
			\midrule 
			{CIFAR-10}&  $\bf{95.04}\pm0.07$ & ${94.11}\pm0.06$ & $0.93$\\
			{CIFAR-100}  &  $\bf{74.45}\pm0.22$ & ${71.89}\pm0.32$ &  $2.56$ \\
			{ImageNet}  &  $\bf{62.08}$ & $59.13$ & $2.95$\\
			\bottomrule
		\end{tabular}
	\end{threeparttable}
	\label{table:steps}
%\end{table}
\end{minipage}
\hspace{2mm}
\begin{minipage}{0.47\linewidth}
\newcommand{\tabincell}[2]{\begin{tabular}{@{}#1@{}}#2\end{tabular}}
%\begin{table} [ht]
	\centering
	\small
	\tabcolsep=0.5mm

    \captionof{table}{Comparison between SLTT and SG on CIFAR-10. ``PG'' means the performance gap between 32 bits and 1 bit.}
	\begin{threeparttable}
		\begin{tabular}{ccccc}
			\toprule  
			Setting & Method & 32-bit & 1-bit & PG \\
			\midrule 
			\multirow{2}*{BL}  &  SLTT & $94.73$ & $94.26$ & ${0.48}$ \\
			&  SG & $\bf{94.85}$ & $\bf{95.16}$ & $-\bf{0.31}$ \\
			\midrule
			{BL+MS}  &  SLTT & $\bf{95.22}$ & $95.04$ & ${0.18}$ \\
			+KL &  SG & ${95.19}$ & $\bf{95.21}$ & $-\bf{0.02}$ \\
			\bottomrule
		\end{tabular}
	\end{threeparttable}
	\label{table:sltt-sg}
%\end{table}
\end{minipage}

\subsection{Comparison Between SLTT and SG}

% \begin{table}[t] 
% 	\caption{\normalsize Comparison between SLTT and SG on CIFAR-10. ``PG'' means the performance gap between 32 bits and 1 bit.}
% 	%\vspace{-7pt}
% 	\label{table:sltt-sg}
% 	\centering
% 	\begin{threeparttable}
% 		\begin{tabular}{ccccc}
% 			\toprule  
% 			Setting & Method & 32-bit & 1-bit & PG \\
% 			\midrule 
% 			\multirow{2}*{BL}  &  SLTT & $94.73$ & $94.26$ & ${0.48}$ \\
% 			&  SG & $\bf{94.85}$ & $\bf{95.16}$ & $-\bf{0.31}$ \\
% 			\midrule
% 			{BL+MS}  &  SLTT & $\bf{95.22}$ & $95.04$ & ${0.18}$ \\
% 			+KL &  SG & ${95.19}$ & $\bf{95.21}$ & $-\bf{0.02}$ \\
% 			\bottomrule
% 		\end{tabular}
% 	\end{threeparttable}
% \end{table}

\vspace{-5pt}

Our new perspective on SNN dynamics was inspired by the SLTT training rule. However, the proposed SEI-BWSNN method is not limited to SLTT, but rather designed for BPTT-based methods, including SLTT, SG, OSTL \citep{bohnstingl2022online}, OTTT \citep{xiao2022online}, and others. 
Specifically, the proposed method considers an SNN (BWSNN) as the self-ensemble of a BANN (BNN), without being tied to any specific training rule. Compared with SLTT, our work additionally introduces a network structure, a distillation technique, and a training pipeline for
BWSNNs. We leverage SLTT to train SNNs due to its superior time and memory efficiency. It should be noticed that even better performance can be achieved with the most commonly used SG method, as shown in \cref{table:sltt-sg}. %We choose to use SLTT for all other experiments as we aim to provide a new perspective on SNN training rather than achieving the best performance.

% \begin{table}[t] 
% 	\caption{\normalsize Comparison between 6 and 1 time steps. The results for CIFAR-10/100 are based on 3 run of experiments. ``PG'' means performance gap between 6 and 1 time steps.}
% 	%\vspace{-7pt}
% 	% \label{table:steps}
% 	\centering
% 	\begin{threeparttable}
% 		\begin{tabular}{l|ccc}
% 			\toprule  
% 			Dataset  & $T=6$ & $T=1$ & PG \\
% 			\midrule 
% 			{CIFAR-10}&  $\bf{95.04}\pm0.07$ & ${94.11}\pm0.06$ & $0.93$\\
% 			{CIFAR-100}  &  $\bf{74.45}\pm0.22$ & ${71.89}\pm0.32$ &  $2.56$ \\
% 			{ImageNet}  &  $\bf{62.08}$ & $59.13$ & $2.95$\\
% 			\bottomrule
% 		\end{tabular}
% 	\end{threeparttable}
% \end{table}

\subsection{Comparison with the State-of-the-Art}
\vspace{-5pt}
\begin{table}[t] \small
	\caption{Comparisons with other SNN training methods. ``TS'' means time steps.}
	\label{table:sota}
	\centering
	\begin{threeparttable}
		\begin{tabular}{c|lcccc}
			\toprule[1.10pt] & Method & Weight & Network & TS  & Mean$\pm$Std \\
			\midrule[1.08pt]
			
			\multirow{6}*{{CIFAR-10}} 
			&Dspike \citep{li2021differentiable} &\multirow{4}*{32-bit}  & ResNet-18  & 6 & $94.25\pm0.07\%$  \\
			& TET \citep{deng2022temporal} &  & ResNet-19 & 6  & $\bf{94.50\pm0.07\%}$  \\
			&SLTT \citep{meng2022towards} & & ResNet-18 & 6  & $94.44\pm0.21\%$ \\
			%&\bf{SEI-BWSNN (ours)} &  & ResNet-18 & 6 & $\bf{95.22}\%$ \\
			\cline{2-6}
            & BitSNNs \cite{hu2024bitsnns} &  \multirow{3}*{1-bit} & ResNet-18 & 1 & 93.74\% \\
            & Shen et al. \cite{shen2024conventional} & & Spikformer-4-384 & 4 & 93.91\% \\
			& \bf{SEI-BWSNN (ours)} &  & ResNet-18 & 6 & $\bf{95.04\pm0.07\%}$ \\
			\midrule[1.08pt]
			
			\multirow{6}*{{CIFAR-100}} 
			&Dspike \citep{li2021differentiable} & \multirow{3}*{32-bit}  & ResNet-18  & 6 & $74.24\pm0.10\%$  \\
			&TET \citep{deng2022temporal} &  & ResNet-19 & 6  & $\bf{74.72\pm0.28\%}$  \\
			&SLTT \citep{meng2022towards} &   & ResNet-18  & 6 & $74.38\pm0.30\%$  \\
			%&\bf{SEI-BWSNN (ours)} &  & ResNet-18 & 6 & $\bf{77.22}\%$ \\
			\cline{2-6}
            & Shen et al. \cite{shen2024conventional} & & Spikformer-4-384 & 4 & 54.54\% \\
			& Wang et al.  \cite{wang2020deep} &  \multirow{3}*{1-bit} & 8-layer CNN & 300 & $62.02\%$ \\
			& \bf{SEI-BWSNN (ours)} &  & ResNet-18 & 6 & $\bf{74.45\pm0.22\%}$ \\
			\midrule[1.08pt]
			
			\multirow{7}*{{ImageNet}} 	
            &Dspike\citep{li2021differentiable} & \multirow{4}*{32-bit}  & ResNet-34  & 6 & ${68.19\%}$  \\
			&TET \citep{deng2022temporal} &   & ResNet-34 & 6  & $64.79\%$ \\
			&SLTT \citep{meng2022towards} &  & NF-ResNet-34 & 6  & ${66.19\%}$ \\
            & QKFormer \cite{zhou2024qkformer} & & QKFormer-10-768 & 4 & $\bf{84.22}\%$ \\
			% &\bf{SEI-BWSNN (ours)} &  & NF-ResNet-34 & 6 & $66.41\%$ \\
			\cline{2-6}
			% & Lu et al. \cite{lu2020exploring} &  \multirow{5}*{1-bit} & VGG-15 & 148 & ${62.71\%}$ \\
            & BESTformer \cite{binaryeventdrivenspikingtransformer} & \multirow{4}*{1-bit} & BESTformer-8-512 & 4 & 63.46\% \\
            & Shen et al. \cite{shen2024conventional} & & Spikformer-8-512 & 1 & 54.54\% \\
            & AGMM \cite{AGMM} & & ResNet-18 & 4 & $64.67\%$ \\
		% & \bf{SEI-BWSNN (ours)} &  & ResNet-34 & 6 & $62.08\%$ \\
            & \bf{SEI-BWSNN (ours)} &  & QKFormer-10-768 & \textbf{2} & $\bf{82.52}\%$ \\

			\midrule[1.08pt]
			
			\multirow{7}*{{DVS-Gesture}} 
			% &STBP-tdBN \citep{zheng2020going} & \multirow{5}*{32-bit} &  ResNet-17 & 40  & $96.87\%$  \\
			& OTTT \citep{xiao2022online} & \multirow{3}*{32-bit}  & VGG-11 & 20   & $96.88\%$  \\	
			&SLTT \citep{meng2022towards} &  & VGG-11 & 10  &  $\mathbf{97.92\pm0.00\%}$  \\
			& PLIF \citep{fang2021incorporating} &   & VGG-like & 20   & $97.57\%$  \\
			%&\bf{SEI-BWSNN (ours)} &   & ResNet-18 & 20 & \underline{97.57\%} \\
			\cline{2-6}
			&Qiao et al. \cite{qiao2021direct} &  \multirow{3}*{1-bit} & 4-layer CNN & 150 & $\bf{97.57\%}$ \\
                % & AGMM \cite{AGMM} & & VGG (8Conv1FC) & 16 & $97.92\%$ \\
                & Pei et al. \cite{pei2023albsnn}& & 5Conv1FC & 20 & 94.63\% \\
			&\bf{SEI-BWSNN (ours)} &   & ResNet-18 & 20 & ${97.11 \pm 0.20\%}$ \\
            \midrule[1.08pt]
			
			\multirow{7}*{{DVS-CIFAR10}} 
			&Dspike\citep{li2021differentiable} & \multirow{3}*{32-bit} &  ResNet-18 & 10  & $75.40\pm0.05\%$  \\
			& TET \citep{deng2022temporal} &   & VGG-11 & 10   & $\mathbf{83.17\pm0.15\%}$  \\	
			&SLTT \citep{meng2022towards} &  & VGG-11 & 10  &  $82.20\pm0.95\%$  \\
			% &\bf{SEI-BWSNN (ours)} &   & ResNet-18 & 10 & \underline{83.00\%} \\
			\cline{2-6}
			% &Qiao et al. \cite{qiao2021direct} &  \multirow{2}*{1-bit} & 4-layer CNN & 30 & $62.10\%$ \\
            & Q-SNN \cite{q-snns} &  \multirow{4}*{1-bit} & VGG (8Conv1FC) & 10 & 81.60\% \\
            & BESTformer \cite{binaryeventdrivenspikingtransformer} &  & BESTformer-2-256 & 10 &78.70\% \\
            % & AGMM \cite{AGMM} & & VGG (8Conv1FC) & 10 & $82.40\%$ \\
            & Shen et al. \cite{shen2024conventional} & & Spikformer-2-256 & 16 & 79.80\% \\
			&\bf{SEI-BWSNN (ours)} &   & ResNet-18 & 10 & $\bf{82.40\pm0.53\%}$ \\
			\bottomrule[1.08pt]
		\end{tabular}
	\end{threeparttable}
    \vspace{-5pt}
\end{table}

This study seeks to rethink SNN dynamics and pave a new path to enhance SNN training. Although we have not explored all avenues of this new path, our proposed method still demonstrates competitive performance compared with the SOTA on both static and neuromorphic datasets, as shown in \cref{table:sota}.
To achieve competitive performance on ImageNet, we adopt the Vision Transformer-based QKFormer \cite{zhou2024qkformer}  architecture as our backbone. We selectively binarize the MLP (FFN) blocks, motivated by their dominant computational footprint in Vision Transformers \cite{lin2022survey}.
Due to the prohibitive cost of training from scratch, we skip both the 
multi-shortcut modification and the first training stage, instead directly fine-tuning the original QKFormer model with KL-divergence loss.
Our BWSNN models with 1-bit weights are compatible with many recent SNN models with 32-bit weights, even when utilizing the same network structures. Compared with other BWSNN models, our approach requires a significantly less number of time steps, and our results are much better on CIFAR-10, CIFAR-100, and DVS-CIFAR10. Furthermore, the performance gap between our BWSNN models and their full-precision counterparts is small, even less than 0.18\% on CIFAR-10 (\cref{table:ablation}), showing that our techniques for training BWSNNs are effective.
Overall, the results in \cref{table:sota} indicate the potential of our proposed SEI-BWSNN method to advance the field of BWSNN training, offering an exciting new direction for further research.

\vspace{-8pt}
\section{Conclusion}
\vspace{-10pt}
In this work, we present a new perspective on the dynamics of SNNs and open up new avenues for training SNNs by leveraging techniques from BNNs. Our analysis of SNN backpropagation reveals that a feedforward SNN can be viewed as a self-ensemble of a binary-activation neural network with noise injection. Based on this perspective, we improve the SNN performance by enhancing the BNN sub-network. Specifically, we employ a structure of multiple shortcuts and a knowledge distillation-based loss function in SNN training. For binary-weight SNNs, we further propose the Self-Ensemble Inspired BWSNN training method (SEI-BWSNN) by incorporating the two-stage training strategy and an update rule for the binary weights, based on the new understanding of the connection between SNNs and BNNs. 
Our method achieves competitive performance for SNNs with both full-precision and binary weights. Notably, for binary-weight SNNs, the obtained model achieves an accuracy of 82.52\% on ImageNet with only 2 time steps, outperforming many current SOTA methods with full-precision weights, showing the great potential of BWSNNs for low-power and low-memory applications.

\bibliographystyle{unsrt}
\bibliography{main}

\begin{thebibliography}{100}

\bibitem{maass1997networks}
Wolfgang Maass.
\newblock Networks of spiking neurons: the third generation of neural network models.
\newblock {\em Neural networks}, 10(9):1659--1671, 1997.

\bibitem{schuman2017survey}
Catherine~D Schuman, Thomas~E Potok, Robert~M Patton, J~Douglas Birdwell, Mark~E Dean, Garrett~S Rose, and James~S Plank.
\newblock A survey of neuromorphic computing and neural networks in hardware.
\newblock {\em arXiv preprint arXiv:1705.06963}, 2017.

\bibitem{neftci2019surrogate}
Emre~O Neftci, Hesham Mostafa, and Friedemann Zenke.
\newblock Surrogate gradient learning in spiking neural networks: Bringing the power of gradient-based optimization to spiking neural networks.
\newblock {\em IEEE Signal Processing Magazine}, 36(6):51--63, 2019.

\bibitem{cramer2022surrogate}
Benjamin Cramer, Sebastian Billaudelle, Simeon Kanya, Aron Leibfried, Andreas Gr{\"u}bl, Vitali Karasenko, Christian Pehle, Korbinian Schreiber, Yannik Stradmann, Johannes Weis, et~al.
\newblock Surrogate gradients for analog neuromorphic computing.
\newblock {\em Proc. Natl. Acad. Sci. {USA}}, 119(4):e2109194119, 2022.

\bibitem{werbos1990backpropagation}
Paul~J Werbos.
\newblock Backpropagation through time: what it does and how to do it.
\newblock {\em Proceedings of the IEEE}, 78(10):1550--1560, 1990.

\bibitem{courbariaux2016binarized}
Matthieu Courbariaux, Itay Hubara, Daniel Soudry, Ran El-Yaniv, and Yoshua Bengio.
\newblock Binarized neural networks: Training deep neural networks with weights and activations constrained to+ 1 or-1.
\newblock {\em arXiv preprint}, arXiv:1602.02830, 2016.

\bibitem{hubara2016binarized}
Itay Hubara, Matthieu Courbariaux, Daniel Soudry, Ran El-Yaniv, and Yoshua Bengio.
\newblock Binarized neural networks.
\newblock In {\em NeurIPS}, volume~29, 2016.

\bibitem{rastegari2016xnor}
Mohammad Rastegari, Vicente Ordonez, Joseph Redmon, and Ali Farhadi.
\newblock Xnor-net: Imagenet classification using binary convolutional neural networks.
\newblock In {\em ECCV}, pages 525--542. Springer, 2016.

\bibitem{hinton2015distilling}
Geoffrey Hinton, Oriol Vinyals, and Jeff Dean.
\newblock Distilling the knowledge in a neural network.
\newblock {\em arXiv preprint arXiv:1503.02531}, 2015.

\bibitem{yan2021near}
Zhanglu Yan, Jun Zhou, and Weng-Fai Wong.
\newblock Near lossless transfer learning for spiking neural networks.
\newblock In {\em AAAI}, 2021.

\bibitem{deng2021optimal}
Shikuang Deng and Shi Gu.
\newblock Optimal conversion of conventional artificial neural networks to spiking neural networks.
\newblock In {\em ICLR}, 2021.

\bibitem{sengupta2019going}
Abhronil Sengupta, Yuting Ye, Robert Wang, Chiao Liu, and Kaushik Roy.
\newblock Going deeper in spiking neural networks: Vgg and residual architectures.
\newblock {\em Frontiers in neuroscience}, 13:95, 2019.

\bibitem{rueckauer2017conversion}
Bodo Rueckauer, Iulia-Alexandra Lungu, Yuhuang Hu, Michael Pfeiffer, and Shih-Chii Liu.
\newblock Conversion of continuous-valued deep networks to efficient event-driven networks for image classification.
\newblock {\em Frontiers in neuroscience}, 11:682, 2017.

\bibitem{han2020rmp}
Bing Han, Gopalakrishnan Srinivasan, and Kaushik Roy.
\newblock {RMP-SNN:} residual membrane potential neuron for enabling deeper high-accuracy and low-latency spiking neural network.
\newblock In {\em CVPR}, 2020.

\bibitem{han2020deep}
Bing Han and Kaushik Roy.
\newblock Deep spiking neural network: Energy efficiency through time based coding.
\newblock In {\em ECCV}, 2020.

\bibitem{ding2021optimal}
Jianhao Ding, Zhaofei Yu, Yonghong Tian, and Tiejun Huang.
\newblock Optimal {ANN-SNN} conversion for fast and accurate inference in deep spiking neural networks.
\newblock In {\em IJCAI}, 2021.

\bibitem{li2021free}
Yuhang Li, Shikuang Deng, Xin Dong, Ruihao Gong, and Shi Gu.
\newblock A free lunch from {ANN}: Towards efficient, accurate spiking neural networks calibration.
\newblock In {\em ICML}, 2021.

\bibitem{meng2022ann}
Qingyan Meng, Shen Yan, Mingqing Xiao, Yisen Wang, Zhouchen Lin, and Zhi-Quan Luo.
\newblock Training much deeper spiking neural networks with a small number of time-steps.
\newblock {\em Neural Networks}, 153:254--268, 2022.

\bibitem{bu2022optimal}
Tong Bu, Wei Fang, Jianhao Ding, Penglin Dai, Zhaofei Yu, and Tiejun Huang.
\newblock Optimal {ANN}-{SNN} conversion for high-accuracy and ultra-low-latency spiking neural networks.
\newblock In {\em ICLR}, 2022.

\bibitem{wang2023toward}
Ziming Wang, Yuhao Zhang, Shuang Lian, Xiaoxin Cui, Rui Yan, and Huajin Tang.
\newblock Toward high-accuracy and low-latency spiking neural networks with two-stage optimization.
\newblock {\em IEEE Transactions on Neural Networks and Learning Systems}, 2023.

\bibitem{zhou2019temporal}
Shibo Zhou, Xiaohua Li, Ying Chen, Sanjeev~T Chandrasekaran, and Arindam Sanyal.
\newblock Temporal-coded deep spiking neural network with easy training and robust performance.
\newblock In {\em AAAI}, 2021.

\bibitem{xiao2021ide}
Mingqing Xiao, Qingyan Meng, Zongpeng Zhang, Yisen Wang, and Zhouchen Lin.
\newblock Training feedback spiking neural networks by implicit differentiation on the equilibrium state.
\newblock In {\em NeurIPS}, 2021.

\bibitem{meng2022training}
Qingyan Meng, Mingqing Xiao, Shen Yan, Yisen Wang, Zhouchen Lin, and Zhi-Quan Luo.
\newblock Training high-performance low-latency spiking neural networks by differentiation on spike representation.
\newblock In {\em CVPR}, 2022.

\bibitem{thiele2019spikegrad}
Johannes~C. Thiele, Olivier Bichler, and Antoine Dupret.
\newblock {SpikeGrad}: An {ANN-equivalent} computation model for implementing backpropagation with spikes.
\newblock In {\em ICLR}, 2020.

\bibitem{wu2021training}
Hao Wu, Yueyi Zhang, Wenming Weng, Yongting Zhang, Zhiwei Xiong, Zheng-Jun Zha, Xiaoyan Sun, and Feng Wu.
\newblock Training spiking neural networks with accumulated spiking flow.
\newblock In {\em AAAI}, 2021.

\bibitem{wu2021tandem}
Jibin Wu, Yansong Chua, Malu Zhang, Guoqi Li, Haizhou Li, and Kay~Chen Tan.
\newblock A tandem learning rule for effective training and rapid inference of deep spiking neural networks.
\newblock {\em IEEE Transactions on Neural Networks and Learning Systems}, 2021.

\bibitem{xiao2022spide}
Mingqing Xiao, Qingyan Meng, Zongpeng Zhang, Yisen Wang, and Zhouchen Lin.
\newblock {SPIDE}: A purely spike-based method for training feedback spiking neural networks.
\newblock {\em Neural Networks}, 161:9--24, 2023.

\bibitem{yang2022training}
Qu~Yang, Jibin Wu, Malu Zhang, Yansong Chua, Xinchao Wang, and Haizhou Li.
\newblock Training spiking neural networks with local tandem learning.
\newblock {\em NeurIPS}, 35:12662--12676, 2022.

\bibitem{yang2023lc}
Qu~Yang, Malu Zhang, Jibin Wu, Kay~Chen Tan, and Haizhou Li.
\newblock {LC-TTFS}: Towards lossless network conversion for spiking neural networks with ttfs coding.
\newblock {\em IEEE Transactions on Cognitive and Developmental Systems}, 2023.

\bibitem{huang2024towards}
Zihan Huang, Xinyu Shi, Zecheng Hao, Tong Bu, Jianhao Ding, Zhaofei Yu, and Tiejun Huang.
\newblock Towards high-performance spiking transformers from {ANN to SNN} conversion.
\newblock In {\em Proceedings of the 32nd ACM International Conference on Multimedia}, pages 10688--10697, 2024.

\bibitem{zenke2021remarkable}
Friedemann Zenke and Tim~P Vogels.
\newblock The remarkable robustness of surrogate gradient learning for instilling complex function in spiking neural networks.
\newblock {\em Neural Computation}, 33(4):899--925, 2021.

\bibitem{wu2018spatio}
Yujie Wu, Lei Deng, Guoqi Li, Jun Zhu, and Luping Shi.
\newblock Spatio-temporal backpropagation for training high-performance spiking neural networks.
\newblock {\em Frontiers in Neuroscience}, 12:331, 2018.

\bibitem{wu2019direct}
Yujie Wu, Lei Deng, Guoqi Li, Jun Zhu, Yuan Xie, and Luping Shi.
\newblock Direct training for spiking neural networks: Faster, larger, better.
\newblock In {\em AAAI}, 2019.

\bibitem{shrestha2018slayer}
Sumit~Bam Shrestha and Garrick Orchard.
\newblock {SLAYER:} spike layer error reassignment in time.
\newblock In {\em NeurIPS}, 2018.

\bibitem{ma2023exploiting}
Gehua Ma, Rui Yan, and Huajin Tang.
\newblock Exploiting noise as a resource for computation and learning in spiking neural networks.
\newblock {\em Patterns}, 4(10), 2023.

\bibitem{zheng2020going}
Hanle Zheng, Yujie Wu, Lei Deng, Yifan Hu, and Guoqi Li.
\newblock Going deeper with directly-trained larger spiking neural networks.
\newblock In {\em AAAI}, 2021.

\bibitem{fang2021sew}
Wei Fang, Zhaofei Yu, Yanqi Chen, Tiejun Huang, Timoth{\'e}e Masquelier, and Yonghong Tian.
\newblock Deep residual learning in spiking neural networks.
\newblock In {\em NeurIPS}, 2021.

\bibitem{li2021differentiable}
Yuhang Li, Yufei Guo, Shanghang Zhang, Shikuang Deng, Yongqing Hai, and Shi Gu.
\newblock Differentiable spike: Rethinking gradient-descent for training spiking neural networks.
\newblock In {\em NeurIPS}, 2021.

\bibitem{guo2022recdis}
Yufei Guo, Xinyi Tong, Yuanpei Chen, Liwen Zhang, Xiaode Liu, Zhe Ma, and Xuhui Huang.
\newblock Recdis-snn: Rectifying membrane potential distribution for directly training spiking neural networks.
\newblock In {\em CVPR}, 2022.

\bibitem{deng2022temporal}
Shikuang Deng, Yuhang Li, Shanghang Zhang, and Shi Gu.
\newblock Temporal efficient training of spiking neural network via gradient re-weighting.
\newblock In {\em ICLR}, 2022.

\bibitem{fang2021incorporating}
Wei Fang, Zhaofei Yu, Yanqi Chen, Timoth{\'e}e Masquelier, Tiejun Huang, and Yonghong Tian.
\newblock Incorporating learnable membrane time constant to enhance learning of spiking neural networks.
\newblock In {\em ICCV}, 2021.

\bibitem{li2022neuromorphic}
Yuhang Li, Youngeun Kim, Hyoungseob Park, Tamar Geller, and Priyadarshini Panda.
\newblock Neuromorphic data augmentation for training spiking neural networks.
\newblock In {\em ECCV}, 2022.

\bibitem{zhang2021rectified}
Malu Zhang, Jiadong Wang, Jibin Wu, Ammar Belatreche, Burin Amornpaisannon, Zhixuan Zhang, Venkata Pavan~Kumar Miriyala, Hong Qu, Yansong Chua, Trevor~E Carlson, et~al.
\newblock Rectified linear postsynaptic potential function for backpropagation in deep spiking neural networks.
\newblock {\em IEEE Transactions on Neural Networks and Learning Systems}, 33(5):1947--1958, 2021.

\bibitem{chowdhurytowards}
Sayeed~Shafayet Chowdhury, Nitin Rathi, and Kaushik Roy.
\newblock Towards ultra low latency spiking neural networks for vision and sequential tasks using temporal pruning.
\newblock In {\em ECCV}, 2022.

\bibitem{guo2022reducing}
Yufei Guo, Yuanpei Chen, Liwen Zhang, YingLei Wang, Xiaode Liu, Xinyi Tong, Yuanyuan Ou, Xuhui Huang, and Zhe Ma.
\newblock Reducing information loss for spiking neural networks.
\newblock In {\em ECCV}, 2022.

\bibitem{yan2024sampling}
Shen Yan, Qingyan Meng, Mingqing Xiao, Yisen Wang, and Zhouchen Lin.
\newblock Sampling complex topology structures for spiking neural networks.
\newblock {\em Neural Networks}, 172:106121, 2024.

\bibitem{meng2022towards}
Qingyan Meng, Mingqing Xiao, Shen Yan, Yisen Wang, Zhouchen Lin, and Zhi-Quan Luo.
\newblock Towards memory- and time-efficient backpropagation for training spiking neural networks.
\newblock In {\em ICCV}, 2023.

\bibitem{xiao2022online}
Mingqing Xiao, Qingyan Meng, Zongpeng Zhang, Di~He, and Zhouchen Lin.
\newblock Online training through time for spiking neural networks.
\newblock In {\em NeurIPS}, 2022.

\bibitem{bellec2020solution}
Guillaume Bellec, Franz Scherr, Anand Subramoney, Elias Hajek, Darjan Salaj, Robert Legenstein, and Wolfgang Maass.
\newblock A solution to the learning dilemma for recurrent networks of spiking neurons.
\newblock {\em Nature Communications}, 11(1):1--15, 2020.

\bibitem{bohnstingl2022online}
Thomas Bohnstingl, Stanis{\l}aw Wo{\'z}niak, Angeliki Pantazi, and Evangelos Eleftheriou.
\newblock Online spatio-temporal learning in deep neural networks.
\newblock {\em IEEE Transactions on Neural Networks and Learning Systems}, 2022.

\bibitem{yin2021accurate}
Bojian Yin, Federico Corradi, and Sander~M Bohte.
\newblock Accurate online training of dynamical spiking neural networks through forward propagation through time.
\newblock {\em arXiv preprint}, arXiv:2112.11231, 2021.

\bibitem{zhang2023self}
Tielin Zhang, Qingyu Wang, and Bo~Xu.
\newblock Self-lateral propagation elevates synaptic modifications in spiking neural networks for the efficient spatial and temporal classification.
\newblock {\em IEEE Transactions on Neural Networks and Learning Systems}, 2023.

\bibitem{liu2018bi}
Zechun Liu, Baoyuan Wu, Wenhan Luo, Xin Yang, Wei Liu, and Kwang-Ting Cheng.
\newblock {Bi-Real Net}: Enhancing the performance of 1-bit {CNNs} with improved representational capability and advanced training algorithm.
\newblock In {\em ECCV}, pages 722--737, 2018.

\bibitem{liu2020reactnet}
Zechun Liu, Zhiqiang Shen, Marios Savvides, and Kwang-Ting Cheng.
\newblock Reactnet: Towards precise binary neural network with generalized activation functions.
\newblock In {\em ECCV}, pages 143--159. Springer, 2020.

\bibitem{martinez2020training}
Brais Martinez, Jing Yang, Adrian Bulat, and Georgios Tzimiropoulos.
\newblock Training binary neural networks with real-to-binary convolutions.
\newblock In {\em ICLR}, 2020.

\bibitem{zhang2022pokebnn}
Yichi Zhang, Zhiru Zhang, and Lukasz Lew.
\newblock {PokeBNN}: A binary pursuit of lightweight accuracy.
\newblock In {\em CVPR}, pages 12475--12485, 2022.

\bibitem{bulat2019improved}
Adrian Bulat, Georgios Tzimiropoulos, Jean Kossaifi, and Maja Pantic.
\newblock Improved training of binary networks for human pose estimation and image recognition.
\newblock {\em arXiv preprint}, arXiv:1904.05868, 2019.

\bibitem{zhu2019binary}
Shilin Zhu, Xin Dong, and Hao Su.
\newblock Binary ensemble neural network: More bits per network or more networks per bit?
\newblock In {\em CVPR}, pages 4923--4932, 2019.

\bibitem{liu2021adam}
Zechun Liu, Zhiqiang Shen, Shichao Li, Koen Helwegen, Dong Huang, and Kwang-Ting Cheng.
\newblock How do adam and training strategies help bnns optimization.
\newblock In {\em ICML}, pages 6936--6946, 2021.

\bibitem{zhuang2018towards}
Bohan Zhuang, Chunhua Shen, Mingkui Tan, Lingqiao Liu, and Ian Reid.
\newblock Towards effective low-bitwidth convolutional neural networks.
\newblock In {\em CVPR}, pages 7920--7928, 2018.

\bibitem{yang2019synetgy}
Yifan Yang, Qijing Huang, Bichen Wu, Tianjun Zhang, Liang Ma, Giulio Gambardella, Michaela Blott, Luciano Lavagno, Kees Vissers, John Wawrzynek, et~al.
\newblock Synetgy: Algorithm-hardware co-design for convnet accelerators on embedded fpgas.
\newblock In {\em Proceedings of the 2019 ACM/SIGDA international symposium on field-programmable gate arrays}, pages 23--32, 2019.

\bibitem{wang2023bitnet}
Hongyu Wang, Shuming Ma, Li~Dong, Shaohan Huang, Huaijie Wang, Lingxiao Ma, Fan Yang, Ruiping Wang, Yi~Wu, and Furu Wei.
\newblock Bitnet: Scaling 1-bit transformers for large language models.
\newblock {\em arXiv preprint arXiv:2310.11453}, 2023.

\bibitem{ma2024era}
Shuming Ma, Hongyu Wang, Lingxiao Ma, Lei Wang, Wenhui Wang, Shaohan Huang, Lifeng Dong, Ruiping Wang, Jilong Xue, and Furu Wei.
\newblock The era of 1-bit {LLMs}: All large language models are in 1.58 bits.
\newblock {\em arXiv preprint arXiv:2402.17764}, 1, 2024.

\bibitem{quantizedspikedriventransformer}
Xuerui Qiu, Malu Zhang, Jieyuan Zhang, Wenjie Wei, Honglin Cao, Junsheng Guo, Rui-Jie Zhu, Yimeng Shan, Yang Yang, and Haizhou Li.
\newblock Quantized spike-driven transformer.
\newblock In {\em ICLR}, 2025.

\bibitem{binaryeventdrivenspikingtransformer}
Honglin Cao, Zijian Zhou, Wenjie Wei, Ammar Belatreche, Yu~Liang, Dehao Zhang, Malu Zhang, Yang Yang, and Haizhou Li.
\newblock Binary event-driven spiking transformer, 2025.

\bibitem{kheradpisheh2022bs4nn}
Saeed~Reza Kheradpisheh, Maryam Mirsadeghi, and Timoth{\'e}e Masquelier.
\newblock Bs4nn: binarized spiking neural networks with temporal coding and learning.
\newblock {\em Neural Processing Letters}, 54(2):1255--1273, 2022.

\bibitem{q-snns}
Wenjie Wei, Yu~Liang, Ammar Belatreche, Yichen Xiao, Honglin Cao, Zhenbang Ren, Guoqing Wang, Malu Zhang, and Yang Yang.
\newblock {Q-SNNs}: Quantized spiking neural networks.
\newblock In {\em Proceedings of the 32nd ACM International Conference on Multimedia}, MM '24, page 8441–8450, 2024.

\bibitem{koo2020sbsnn}
Minsuk Koo, Gopalakrishnan Srinivasan, Yong Shim, and Kaushik Roy.
\newblock Sbsnn: Stochastic-bits enabled binary spiking neural network with on-chip learning for energy efficient neuromorphic computing at the edge.
\newblock {\em IEEE Transactions on Circuits and Systems I: Regular Papers}, 67(8):2546--2555, 2020.

\bibitem{srinivasan2019restocnet}
Gopalakrishnan Srinivasan and Kaushik Roy.
\newblock {ReStoCNet}: Residual stochastic binary convolutional spiking neural network for memory-efficient neuromorphic computing.
\newblock {\em Frontiers in neuroscience}, 13:189, 2019.

\bibitem{hu2021quantized}
SG~Hu, GC~Qiao, TP~Chen, Qi~Yu, Y~Liu, and LM~Rong.
\newblock Quantized {STDP-based} online-learning spiking neural network.
\newblock {\em Neural Computing and Applications}, 33(19):12317--12332, 2021.

\bibitem{qiao2021direct}
GC~Qiao, Ning Ning, Y~Zuo, SG~Hu, Qi~Yu, and Y~Liu.
\newblock Direct training of hardware-friendly weight binarized spiking neural network with surrogate gradient learning towards spatio-temporal event-based dynamic data recognition.
\newblock {\em Neurocomputing}, 457:203--213, 2021.

\bibitem{wang2020deep}
Yixuan Wang, Yang Xu, Rui Yan, and Huajin Tang.
\newblock Deep spiking neural networks with binary weights for object recognition.
\newblock {\em IEEE Transactions on Cognitive and Developmental Systems}, 13(3):514--523, 2020.

\bibitem{EsserMACAABMMBN16}
Steven~K. Esser, Paul~A. Merolla, John~V. Arthur, Andrew~S. Cassidy, Rathinakumar Appuswamy, Alexander Andreopoulos, David~J. Berg, Jeffrey~L. McKinstry, Timothy Melano, Davis~R. Barch, Carmelo di~Nolfo, Pallab Datta, Arnon Amir, Brian Taba, Myron~D. Flickner, and Dharmendra~S. Modha.
\newblock Convolutional networks for fast, energy-efficient neuromorphic computing.
\newblock {\em Proc. Natl. Acad. Sci. {USA}}, 113(41):11441--11446, 2016.

\bibitem{wei2021binarized}
Ming-Liang Wei, Mikail Yayla, Shu-Yin Ho, Jian-Jia Chen, Chia-Lin Yang, and Hussam Amrouch.
\newblock Binarized snns: Efficient and error-resilient spiking neural networks through binarization.
\newblock In {\em 2021 IEEE/ACM International Conference On Computer Aided Design (ICCAD)}, pages 1--9. IEEE, 2021.

\bibitem{qp-snns}
Wenjie Wei, Malu Zhang, Zijian Zhou, Ammar Belatreche, Yimeng Shan, Yu~Liang, Honglin Cao, Jieyuan Zhang, and Yang Yang.
\newblock {QP-SNN}: Quantized and pruned spiking neural networks.
\newblock In {\em ICLR}, 2025.

\bibitem{lu2020exploring}
Sen Lu and Abhronil Sengupta.
\newblock Exploring the connection between binary and spiking neural networks.
\newblock {\em Frontiers in neuroscience}, 14:535, 2020.

\bibitem{AGMM}
Yu~Liang, Wenjie Wei, Ammar Belatreche, Honglin Cao, Zijian Zhou, Shuai Wang, Malu Zhang, and Yang Yang.
\newblock Towards accurate binary spiking neural networks: Learning with adaptive gradient modulation mechanism.
\newblock In {\em AAAI}, volume~39, pages 1402--1410, 2025.

\bibitem{bengio2013estimating}
Yoshua Bengio, Nicholas L{\'e}onard, and Aaron Courville.
\newblock Estimating or propagating gradients through stochastic neurons for conditional computation.
\newblock {\em arXiv preprint}, arXiv:1308.3432, 2013.

\bibitem{wang2020sparsity}
Peisong Wang, Xiangyu He, Gang Li, Tianli Zhao, and Jian Cheng.
\newblock Sparsity-inducing binarized neural networks.
\newblock In {\em AAAI}, volume~34, pages 12192--12199, 2020.

\bibitem{tu2022adabin}
Zhijun Tu, Xinghao Chen, Pengju Ren, and Yunhe Wang.
\newblock Adabin: Improving binary neural networks with adaptive binary sets.
\newblock In {\em ECCV}, pages 379--395. Springer, 2022.

\bibitem{noh2017regularizing}
Hyeonwoo Noh, Tackgeun You, Jonghwan Mun, and Bohyung Han.
\newblock Regularizing deep neural networks by noise: Its interpretation and optimization.
\newblock In {\em NeurIPS}, volume~30, 2017.

\bibitem{liu2018towards}
Xuanqing Liu, Minhao Cheng, Huan Zhang, and Cho-Jui Hsieh.
\newblock Towards robust neural networks via random self-ensemble.
\newblock In {\em ECCV}, pages 369--385, 2018.

\bibitem{wang2019resnets}
Bao Wang, Zuoqiang Shi, and Stanley Osher.
\newblock {ResNets} ensemble via the {Feynman-Kac} formalism to improve natural and robust accuracies.
\newblock In {\em NeurIPS}, volume~32, 2019.

\bibitem{bethge2021meliusnet}
Joseph Bethge, Christian Bartz, Haojin Yang, Ying Chen, and Christoph Meinel.
\newblock Meliusnet: An improved network architecture for binary neural networks.
\newblock In {\em Proceedings of the IEEE/CVF Winter Conference on Applications of Computer Vision}, pages 1439--1448, 2021.

\bibitem{he2016deep}
Kaiming He, Xiangyu Zhang, Shaoqing Ren, and Jian Sun.
\newblock Deep residual learning for image recognition.
\newblock In {\em CVPR}, 2016.

\bibitem{he2016identity}
Kaiming He, Xiangyu Zhang, Shaoqing Ren, and Jian Sun.
\newblock Identity mappings in deep residual networks.
\newblock In {\em ECCV}, 2016.

\bibitem{kundu2021spike}
Souvik Kundu, Gourav Datta, Massoud Pedram, and Peter~A Beerel.
\newblock Spike-thrift: Towards energy-efficient deep spiking neural networks by limiting spiking activity via attention-guided compression.
\newblock In {\em Proceedings of the IEEE/CVF Winter Conference on Applications of Computer Vision}, pages 3953--3962, 2021.

\bibitem{davies2018loihi}
Mike Davies, Narayan Srinivasa, Tsung-Han Lin, Gautham Chinya, Yongqiang Cao, Sri~Harsha Choday, Georgios Dimou, Prasad Joshi, Nabil Imam, Shweta Jain, et~al.
\newblock Loihi: A neuromorphic manycore processor with on-chip learning.
\newblock {\em IEEE Micro}, 38(1):82--99, 2018.

\bibitem{zagoruyko2016wide}
Sergey Zagoruyko and Nikos Komodakis.
\newblock Wide residual networks.
\newblock {\em arXiv preprint}, arXiv:1605.07146, 2016.

\bibitem{krizhevsky2009learning}
Alex Krizhevsky and Geoffrey Hinton.
\newblock Learning multiple layers of features from tiny images.
\newblock 2009.

\bibitem{deng2009imagenet}
Jia Deng, Wei Dong, Richard Socher, Li-Jia Li, Kai Li, and Li~Fei-Fei.
\newblock {ImageNet}: A large-scale hierarchical image database.
\newblock In {\em CVPR}, 2009.

\bibitem{amir2017low}
Arnon Amir, Brian Taba, David Berg, Timothy Melano, Jeffrey McKinstry, Carmelo Di~Nolfo, Tapan Nayak, Alexander Andreopoulos, Guillaume Garreau, Marcela Mendoza, Jeff Kusnitz, Michael Debole, Steve Esser, Tobi Delbruck, Myronc Flickner, and Dharmendra Modha.
\newblock A low power, fully event-based gesture recognition system.
\newblock In {\em CVPR}, 2017.

\bibitem{li2017cifar10}
Hongmin Li, Hanchao Liu, Xiangyang Ji, Guoqi Li, and Luping Shi.
\newblock {CIFAR10-DVS}: an event-stream dataset for object classification.
\newblock {\em Frontiers in neuroscience}, 11:309, 2017.

\bibitem{brock2021high}
Andy Brock, Soham De, Samuel~L Smith, and Karen Simonyan.
\newblock High-performance large-scale image recognition without normalization.
\newblock In {\em ICML}, pages 1059--1071, 2021.

\bibitem{chen2021bnn}
Tianlong Chen, Zhenyu Zhang, Xu~Ouyang, Zechun Liu, Zhiqiang Shen, and Zhangyang Wang.
\newblock " {BNN-BN}=?": Training binary neural networks without batch normalization.
\newblock In {\em CVPR}, pages 4619--4629, 2021.

\bibitem{he2015delving}
Kaiming He, Xiangyu Zhang, Shaoqing Ren, and Jian Sun.
\newblock Delving deep into rectifiers: Surpassing human-level performance on imagenet classification.
\newblock In {\em ICCV}, pages 1026--1034, 2015.

\bibitem{hu2018squeeze}
Jie Hu, Li~Shen, and Gang Sun.
\newblock Squeeze-and-excitation networks.
\newblock In {\em CVPR}, pages 7132--7141, 2018.

\bibitem{hu2024bitsnns}
Yangfan Hu, Qian Zheng, and Gang Pan.
\newblock Bitsnns: Revisiting energy-efficient spiking neural networks.
\newblock {\em IEEE Transactions on Cognitive and Developmental Systems}, 2024.

\bibitem{shen2024conventional}
Guobin Shen, Dongcheng Zhao, Tenglong Li, Jindong Li, and Yi~Zeng.
\newblock Are conventional {SNNs} really efficient? a perspective from network quantization.
\newblock In {\em CVPR}, pages 27538--27547, 2024.

\bibitem{zhou2024qkformer}
Chenlin Zhou, Han Zhang, Zhaokun Zhou, Liutao Yu, Liwei Huang, Xiaopeng Fan, Li~Yuan, Zhengyu Ma, Huihui Zhou, and Yonghong Tian.
\newblock {QKF}ormer: Hierarchical spiking transformer using {Q-K} attention.
\newblock In {\em NeurIPS}, 2024.

\bibitem{pei2023albsnn}
Yijian Pei, Changqing Xu, Zili Wu, Yi~Liu, and Yintang Yang.
\newblock {ALBSNN}: ultra-low latency adaptive local binary spiking neural network with accuracy loss estimator.
\newblock {\em Frontiers in Neuroscience}, Volume 17 - 2023, 2023.

\bibitem{lin2022survey}
Tianyang Lin, Yuxin Wang, Xiangyang Liu, and Xipeng Qiu.
\newblock A survey of transformers.
\newblock {\em AI open}, 3:111--132, 2022.

\bibitem{devries2017improved}
Terrance DeVries and Graham~W Taylor.
\newblock Improved regularization of convolutional neural networks with cutout.
\newblock {\em arXiv preprint arXiv:1708.04552}, 2017.

\bibitem{rathi2020diet}
Nitin Rathi and Kaushik Roy.
\newblock {DIET-SNN}: A low-latency spiking neural network with direct input encoding and leakage and threshold optimization.
\newblock {\em IEEE Transactions on Neural Networks and Learning Systems}, 2021.

\bibitem{SpikingJelly}
Wei Fang, Yanqi Chen, Jianhao Ding, Zhaofei Yu, Timothée Masquelier, Ding Chen, Liwei Huang, Huihui Zhou, Guoqi Li, and Yonghong Tian.
\newblock Spikingjelly: An open-source machine learning infrastructure platform for spike-based intelligence.
\newblock {\em Science Advances}, 9(40):eadi1480, 2023.

\bibitem{brock2021characterizing}
Andrew Brock, Soham De, and Samuel~L Smith.
\newblock Characterizing signal propagation to close the performance gap in unnormalized {ResNets}.
\newblock In {\em ICLR}, 2021.

\bibitem{qiao2019micro}
Siyuan Qiao, Huiyu Wang, Chenxi Liu, Wei Shen, and Alan Yuille.
\newblock Micro-batch training with batch-channel normalization and weight standardization.
\newblock {\em arXiv preprint arXiv:1903.10520}, 2019.

\bibitem{paszke2019pytorch}
Adam Paszke, Sam Gross, Francisco Massa, Adam Lerer, James Bradbury, Gregory Chanan, Trevor Killeen, Zeming Lin, Natalia Gimelshein, Luca Antiga, et~al.
\newblock {PyTorch}: An imperative style, high-performance deep learning library.
\newblock In {\em NeurIPS}, 2019.

\bibitem{rumelhart1986learning}
David~E Rumelhart, Geoffrey~E Hinton, and Ronald~J Williams.
\newblock Learning representations by back-propagating errors.
\newblock {\em Nature}, 323(6088):533--536, 1986.

\bibitem{kingmaadam}
Diederick~P Kingma and Jimmy Ba.
\newblock Adam: A method for stochastic optimization.
\newblock In {\em ICLR}, 2015.

\bibitem{loshchilov2016sgdr}
Ilya Loshchilov and Frank Hutter.
\newblock {SGDR}: Stochastic gradient descent with warm restarts.
\newblock In {\em ICLR}, 2017.

\end{thebibliography}

%%%%%%%%%%%%%%%%%%%%%%%%%%%%%%%%%%%%%%%%%%%%%%%%%%%%%%%%%%%%

\appendix

\section{Dataset Description and Preprocessing}

\paragraph{CIFAR-10} The CIFAR-10 dataset~\citep{krizhevsky2009learning} is a collection of 60,000 32$\times$32 color images distributed across 10 distinct classes, with 50,000 training images and 10,000 testing images. To ensure uniformity, we normalize the image data to achieve zero mean and unit variance. To augment the data, we randomly crop each image by 4 pixels on each border, flip it horizontally, and apply cutout \citep{devries2017improved}. In addition, we employ direct encoding \citep{rathi2020diet} to encode the image pixels into time series. This encoding entails repeatedly presenting the pixel values to the input layer at each time step. CIFAR-10 is licensed under MIT.

\paragraph{CIFAR-100} The CIFAR-100 dataset~\citep{krizhevsky2009learning} is comparable to CIFAR-10, with the sole difference being that it encompasses 100 object classes. The dataset comprises 50,000 training samples and 10,000 testing samples, with each image being a 32$\times$32 color image. We use the same data preprocessing and input encoding techniques as we do for CIFAR-10. CIFAR-100 is licensed under MIT.

\paragraph{ImageNet} The ImageNet-1K dataset \citep{deng2009imagenet} comprises color images that are distributed among 1000 object classes, with 1,281,167 training images and 50,000 validation images. This dataset is licensed under the Custom (non-commercial) license. To achieve uniformity, we normalize the image data to obtain zero mean and unit variance. For training samples, we randomly resize the images and crop them to size 224$\times$224, following which we apply horizontal flipping. For validation samples, we resize the images to 256$\times$256 and center-crop them to 224$\times$224. We also employ direct encoding \citep{rathi2020diet} to transform the pixels into time sequences, similar to CIFAR-10 and CIFAR-100.

\paragraph{DVS-CIFAR10} The DVS-CIFAR10 dataset \citep{li2017cifar10} is a neuromorphic dataset created by converting CIFAR-10 using a DVS camera. It comprises 10,000 event-based images with expanded pixel dimensions of 128$\times$128. The dataset is licensed under CC BY 4.0. We partition the dataset into 9000 training images and 1000 testing images. As far as data preprocessing is concerned, we incorporate the events into frames \citep{fang2021incorporating} and interpolate to reduce the spatial resolution to 48$\times$48. For data augmentation, we use random horizontal flipping and random rolling within 5 pixels, following the method described in \citep{deng2022temporal}.

\paragraph{DVS-Gesture} 
The DVS-Gesture dataset \citep{amir2017low} is captured using a Dynamic Vision Sensor (DVS) and comprises spike trains with two channels representing ON- and OFF-event spikes. It includes 11 hand gestures from 29 subjects under three illumination conditions, with 1176 training samples and 288 testing samples. The dataset is licensed under Creative Commons Attribution 4.0. For data preprocessing, we adopt the method proposed in \citep{fang2021incorporating} to integrate the events into frames, leveraging the SpikingJelly \citep{SpikingJelly} framework.

\section{Network Architectures}

\subsection{Normalization-Free ResNet}
\label{sec:nfnet}
In our study, we propose a novel perspective on the dynamics of a feedforward SNN by treating it as a self-ensemble of a binary-activation neural network (BANN). This new perspective implies to treat the BANN sub-networks at each time step as independent. As a result, we encounter a challenge in adopting the effective technique of batch normalization (BN) along the temporal dimension \citep{zheng2020going,li2021differentiable,deng2022temporal,meng2022training} for ImageNet, since it requires gathering data from all time steps to calculate the mean and variance statistics. To address this challenge, we turn to the normalization-free ResNet (NF-ResNet) \citep{brock2021characterizing,brock2021high}, which is a BN-free ResNet structure.

The NF-ResNet structure replaces BN with scaled weight standardization (sWS) \citep{brock2021characterizing,qiao2019micro}, which normalizes the weights according to:
\begin{equation}
\hat{\mathbf{W}}_{i, j} = \gamma \frac{\mathbf{W}_{i, j}-\mu_{\mathbf{W}_{i, \cdot}}}{\sigma_{\mathbf{W}_{i, \cdot}}},
\end{equation}
where the mean $\mu_{\mathbf{W}_{i, \cdot}}$ and standard deviation $\sigma_{\mathbf{W}_{i, \cdot}}$ are computed across the fan-in extent indexed by $i$, and $N$ is the dimension of the fan-in extent. The hyperparameter $\gamma$ is introduced to stabilize the signal propagation during the forward pass. To determine $\gamma$, consider a single layer $\mathbf{z}=\hat{\mathbf{W}}g(\mathbf{x})$, where $g(\cdot)$ is the activation and $\mathbf{x}$ is a vector of independent and identically distributed (i.i.d.) elements with a normal distribution of mean 0 and standard deviation 1.
We determine the value of $\gamma$ that ensures that the expected value of $\mathbf{z}$ is 0, and the covariance of $\mathbf{z}$ is the identity matrix $\mathbf{I}$. 
For SNNs, we can represent the activation $g(\cdot)$ at each time step using the Heaviside step function. Then we set $\gamma\approx2.74$, as calculated by \citep{xiao2022online}, to ensure stable forward propagation.
Additionally, the sWS component incorporates a learnable scaling factor for the weights to mimic the scaling factor of BN. SWS does not introduce any dependence on data from different batches and time steps, making it a compelling alternative for replacing BN in our framework.

The signal-preserving property of sWS in deep ResNets is not as effective as BN due to the presence of skip connections. To address this, the NF-ResNet structure incorporates an additional technique to preserve signal during the forward pass. Specifically, NF-ResNet employs residual blocks of the form $x_{l+1}=x_{l}+\alpha f_{l}\left(x_{l} / \beta_{l}\right)$, where $\alpha$ and $\beta_l$ are hyperparameters used to stabilize signals, and sWS is imposed on the weights in $f_l(\cdot)$. The structure is illustrated in \cref{fig:nfblock}.
To ensure that $f_l(x_l/\beta_l)$ has unit variance, carefully chosen $\beta_l$ values and sWS components are used together. The parameter $\alpha$ controls the rate of variance growth between blocks and is set to 0.2 in our experiments. For more information on the network design, please see \citep{brock2021characterizing}.

\subsection{Summary of Used Network Architectures}

\begin{figure}[t]
	%\centering
	\begin{subfigure}{0.49\linewidth}
		\centering
		\includegraphics[height=0.99\columnwidth]{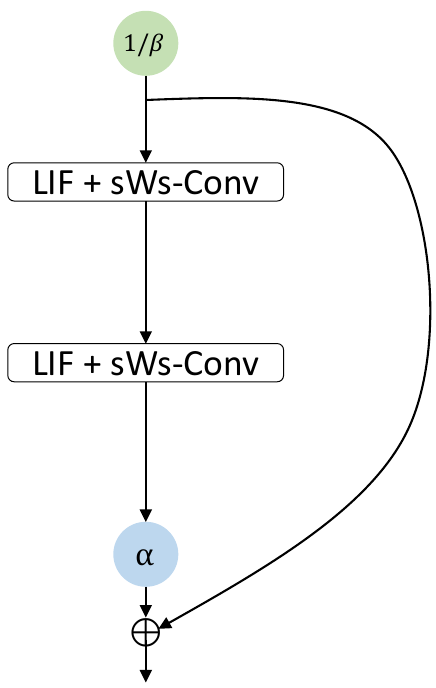}
		\caption{} 
		\label{fig:nfblock}
	\end{subfigure}
	\begin{subfigure}{0.49\linewidth}
		\centering
		\includegraphics[height=0.99\columnwidth]{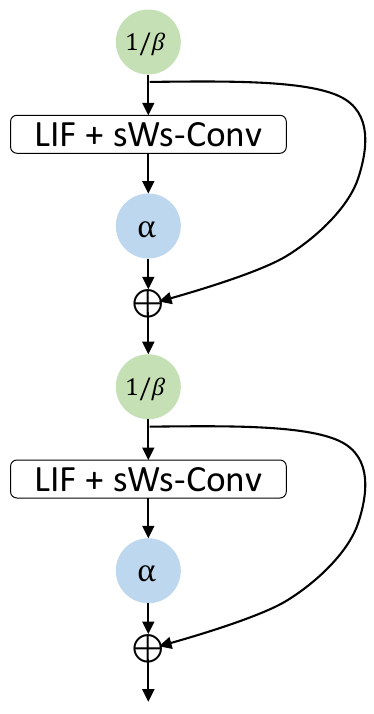}
		\caption{}  
		\label{fig:bnfblock}
	\end{subfigure}
	%\vspace{-7pt}
	\caption{(a) The original basic block structure in a spiking NF-ResNet. (b) The proposed basic block structure with double skip connections in a spiking NF-ResNet.}
	%\vspace{-12pt}
\end{figure}

We conduct experiments on CIFAR-10, CIFAR-100, and DVS-CIFAR10 using ResNet-18 \citep{he2016deep} with pre-activation residual blocks \citep{he2016identity}. The structure of multiple shortcuts is equipped in the networks.  All ReLU activations are substituted with leaky integrate and fire (LIF) neurons to make the network implementable for neuromorphic computing. Additionally, we replace all max pooling operations with average pooling. BN components calculate the mean and variance statistics for each time step rather than the total time horizon. Specifically, in each iteration, a BN component is implemented for $T$ times, where $T$ represents the total number of time steps.
For experiments on ImageNet, we employ NF-ResNet-34 with multiple shortcuts, as shown in \cref{fig:bnfblock}.

\section{Training Settings}

The implementation relies on the PyTorch framework \citep{paszke2019pytorch} and SpikingJelly \citep{SpikingJelly} toolkit, while the experiments are conducted using either a Tesla-V100 or a Tesla-A100 GPU.

To enhance the performance, we incorporate a regularizer suggested in Deng et al. \cite{deng2022temporal} into the loss function, resulting in
\begin{equation}  \label{eqn:appendix_loss}
\begin{aligned}
\mathcal{L}_{\text {SNN}}  =  -\frac{1}{nT} \sum_{t=1}^T  \sum_{i=1}^n  
\Bigg[\left(1-\lambda\right)\left( \sum_c \rho_c^{\mathcal{A}}\left(X_i[t]\right)  \log \left(\frac{\rho_c^{\mathcal{S}}\left(X_i\right)[t]}{\rho_c^{\mathcal{A}}\left(X_i[t]\right)}\right)\right) 
 + \lambda ||{O}_i[t]-V_{th}||^2  \Bigg] ,
\end{aligned}
\end{equation}
where $T$ is the number of total time steps, $c$ represents classes, $n$ denotes the batch size, $X_i$ is the time sequence input data with $X_i[t]$ being the data at the $t$-th time step, ${O}_i[t]$ is the network output at the $t$-th time step, $\rho_c^{\mathcal{A}}$ and $\rho_c^{\mathcal{S}}$ are the softmax outputs of the full-precision ANN teacher network and (binary) SNN student network, respectively, $V_{th}$ is the spike threshold, and $\lambda$ is a hyperparameter chosen differently for each dataset. For CIFAR-10, CIFAR-100, ImageNet, and DVS-CIFAR10, $\lambda$ is set to be 0, 0.05, 0.001, and 0.001, respectively.

\begin{table}[t]
	\caption{Training hyperparameters for stage-1 and stage-2. ``$T$'' means the number of time steps, ``Opt'' means the used optimizer, ``LR'' means learning rate, ``BS'' means batch size, ``WD'' means weight decay, ``WRN'' means the wide ResNet architecture, and ``RN'' means the ResNet architecture.}
	\label{table:parameter}
	\centering
	\begin{tabular}{cccccc}
		\hline &  &  CIFAR-10 & CIFAR-100 & ImageNet & DVS-CIFAR10 \\
		\hline \multirow{7}*{\rotatebox{90}{Stage-1}}  & Epoch & 300 & 300 & 100 & 300 \\
            & T & 6 & 6 & 1 & 10 \\
            & Opt & SGD & SGD & Adam & Adam \\
            & LR & 0.1 & 0.1 & 0.001 & 0.001 \\
            & BS & 128 & 128 & 512 & 128 \\
            & WD & $5\times10^{-5}$ & $5\times10^{-4}$ & $1\times10^{-5}$ & 0 \\
            & Teacher & WRN-28$\times$10 & WRN-28$\times$10 & RN-50 & / \\
            \hline \multirow{8}*{\rotatebox{90}{Stage-2}}  & \multirow{2}*{Epoch} & \multirow{2}*{300} & \multirow{2}*{300} & 30 (32-bit) & \multirow{2}*{300} \\
            &  &  &  & 100 (1-bit) & \\
            & T & 6 & 6 & 6 & 10 \\
            & Opt & Adam & Adam & Adam & Adam \\
            & LR & 0.001 & 0.001 & 0.0001 & 0.001 \\
            & BS & 128 & 128 & 512 & 128 \\
            & WD & 0 & 0 & 0 & 0 \\
            & Teacher & WRN-28$\times$10 & WRN-28$\times$10 & RN-50 & / \\
		\hline
	\end{tabular}
\end{table}

For all the tasks, we use SGD \citep{rumelhart1986learning} with momentum 0.9, or use Adam \citep{kingmaadam} with the hyperpatameter $\beta_{1}$ and $\beta_{2}$ set to be 0.9 and 0.999, respectively. We use cosine annealing \citep{loshchilov2016sgdr} as the learning rate schedule. Our teacher models are either WideResNet-28$\times$10 \citep{zagoruyko2016wide} or ResNet-50 \citep{he2016deep}.
For CIFAR-10, CIFAR-100, and DVS-CIFAR10, we conduct two-stage training where we first train full-precision SNNs and then use the obtained models as initialization to train binary SNNs.
For ImageNet, the two-stage training strategy is slightly different. We train full-precision SNNs with 1 time step in the first stage, and then use the obtained models as initialization to train both full-precision and binary SNNs with 6 time steps.
Detailed hyperparameters and settings are provided in \cref{table:parameter}. 

\section{Additional Experiments}

\subsection{Overhead of the Knowledge Distillation Method}
In this work, we apply a knowledge distillation (KD) technique, the KL divergence loss, to improve performance. While KD yields a notable 4\% improvement for certain datasets (refer to \cref{table:ablation}), the overhead is affordable. Firstly, our teacher ANN models are sourced directly from open-source projects, eliminating the need for tailoring to SNN models. Secondly, the computation of the KL loss does not significantly increase training overhead, as demonstrated in \cref{table:kl_overhead} (the first two elements in each triad). Notably, despite the overhead introduced by KD, our approach remains more memory- and time-efficient than the commonly used BPTT training pipeline, owing to our adoption of SLTT training \cite{meng2022towards}.

\begin{table}[t] 
        \caption{Comparison of training memory cost and training time between different settings related to KD. The triad in each cell represents the values of SLTT training with KD technique (our method), SLTT training without KD technique, and BPTT training without KD technique, respectively.}
        \label{table:kl_overhead}
    \centering
	\begin{threeparttable}
		\begin{tabular}{cccc}
			\toprule  
			& CIFAR-10  & CIFAR-100 & ImageNet \\
			\midrule 
			Memory & {1.8G, 1.6G, 4.4G} &  1.8G, 1.5G, 4.4G  & 15.3G, 14.3G, OOM\\
			Time/epoch & {73s, \ 62s, \ \ 83s}  &  {104s, \ 92s, \ 125s} & {3.0h, \ \  2.9h, \ \ \ -}  \\
			\bottomrule
		\end{tabular}
	\end{threeparttable}
\end{table}

\subsection{Smaller Latency Setting}
Recent works on SNN training typically set the number of time steps to be 4 or 6 on static datasets, while we only conduct experiments with $T=6$. With $T=4$, our method can still achineve good accuracies, quite similar to the results for $T=6$, as shown in \cref{table:timestep4}.

\begin{table}[t] 
        \caption{Performance for different time steps on CIFAR-10 and CIFAR-100.}
        \label{table:timestep4}
        \centering
	\begin{threeparttable}
		\begin{tabular}{cccc}
			\toprule  
			& CIFAR-10  & CIFAR-100 \\
			\midrule 
			T = 4 & 94.98\% &  74.38\%  \\
			T = 6 & 95.04\%  &  74.45\%  \\
			\bottomrule
		\end{tabular}
	\end{threeparttable}
\end{table}

\section{Broader Impact and Limitation}

There is no direct negative broader impact since this work focuses on effective training methods for binary-weight SNNs. Our work emphasizes the direct benefits that BNN research can offer to the SNN community. This finding is novel and can further inspire simple and undiscovered methods for improving SNN such as channel-wise scaling. Furthermore, our pipeline creates strong binary-weight SNNs that can outperform many SOTA models relying on full-precision weights, highlighting the potential of BWSNN as a promising approach for ultra-low power consumption.

As for the limitation, since we use SLTT for training, our work inherits the same limitations, such as the inability to integrate certain network techniques like temporal dimension BN due to the online training nature. This prompts further exploration into techniques compatible with the online learning paradigm of SNNs.

\end{document}